\documentclass[journal]{IEEEtran}

\usepackage{graphicx}
\usepackage{color}
 \usepackage[table,xcdraw]{xcolor}

\usepackage{comment}
\usepackage{epsfig}
\usepackage{graphicx}
\usepackage{amsmath}
\usepackage{amssymb}
\usepackage{comment}
\usepackage{multirow}

\usepackage{hyperref}

\usepackage{booktabs}
\usepackage{array, caption, threeparttable}
\usepackage{caption}
\usepackage{subfigure}
\usepackage{algorithm}
\usepackage{listings}
\usepackage{color}
\usepackage{mathtools}

\usepackage{todonotes}
\usepackage{gensymb}
\usepackage{multirow}

\DeclareUnicodeCharacter{2212}{-}

\begin{document}

\title{Benchmarking Joint Face Spoofing and Forgery Detection with Visual and Physiological Cues}

\author{Zitong Yu, Rizhao Cai, Zhi Li, Wenhan Yang, Jingang Shi and Alex C. Kot,~\IEEEmembership{Fellow,~IEEE}


\thanks{Z. Yu, R. Cai, Z. Li, W. Yang and A. Kot are with ROSE Lab, Nanyang Technological University, Singapore. E-mail: \{zitong.yu, rzcai, zhi003, wenhan.yang, eackot\}@ntu.edu.sg}

\thanks{J. Shi is with Xi'an Jiaotong University, China. E-mail: jingang@xjtu.edu.cn}}

\markboth{IEEE Transactions on Information Forensics and Security}%
{Shell \MakeLowercase{\textit{et al.}}: Bare Demo of IEEEtran.cls for IEEE Journals}

\maketitle

\begin{abstract}
Face anti-spoofing (FAS) and face forgery detection play vital roles in securing face biometric systems from presentation attacks (PAs) and vicious digital manipulation (e.g., deepfakes). Despite promising performance upon large-scale data and powerful deep models, the generalization problem of existing approaches is still an open issue. Most of recent approaches focus on 1) unimodal visual appearance or physiological (i.e., remote photoplethysmography (rPPG)) cues; and 2) separated feature representation for FAS or face forgery detection. On one side, unimodal appearance and rPPG features are respectively vulnerable to high-fidelity face 3D mask and video replay attacks, inspiring us to design reliable multi-modal fusion mechanisms for generalized face attack detection. On the other side, there are rich common features across FAS and face forgery detection tasks (e.g., periodic rPPG rhythms and vanilla appearance for bonafides), providing solid evidence to design a joint FAS and face forgery detection system in a multi-task learning fashion. 
In this paper, we establish the first joint face spoofing and forgery detection benchmark using both visual appearance and physiological rPPG cues. To enhance the rPPG periodicity discrimination, we design a two-branch physiological network using both facial spatio-temporal rPPG signal map and its continuous wavelet transformed counterpart as inputs. To mitigate the modality bias and improve the fusion efficacy, we conduct a weighted batch and layer normalization for both appearance and rPPG features before multi-modal fusion. We also investigate prevalent deep models, feature fusion strategies and multi-task learning configurations for joint face spoofing and forgery detection. We find that the generalization capacities of both unimodal (appearance or rPPG) and multi-modal (appearance+rPPG) models can be obviously improved via joint training on these two tasks. We hope this new benchmark will facilitate the future research of both FAS and deepfake detection communities. 

\end{abstract}

\begin{IEEEkeywords}
Face anti-spoofing, face forgery detection, deepfake, rPPG, fusion, joint training, multi-task learning.
\end{IEEEkeywords}

\IEEEpeerreviewmaketitle

\section{Introduction}


\IEEEPARstart{D}{ue} to its convenience and remarkable accuracy, face recognition technology has been applied in a few interactive intelligent applications such as checking-in and mobile payment. However, existing face recognition systems are vulnerable to presentation attacks (PAs) ranging from print, replay, makeup, 3D-mask, etc. Therefore, both academia and industry have paid extensive attention to developing face anti-spoofing (FAS)~\cite{yu2021deep} technology for securing the face recognition system. 
In the early stage, according to the bonafide/spoof discrepancy on facial appearance~\cite{galbally2014face} (e.g., high visual quality for the bonafide vs. quality distortion with artifacts for the spoof) and physiological dynamics~\cite{li2016generalized} (e.g., periodic heartbeat rhythm for the bonafide vs. noisy remote photoplethysmography (rPPG)~\cite{yu2021facial} signals for the spoof), plenty of traditional handcrafted visual texture/quality features~\cite{Komulainen2014Context,Patel2016Secure} and physiological liveness cues~\cite{li2016generalized,lin2019face,liu20163d,liu2021multi} have been proposed for FAS. Subsequently, thanks to the emergence of large-scale public FAS datasets~\cite{zhang2020celeba,liu2019deep,george2019biometric,liu2022contrastive} with rich attack types and recorded sensors, more and more deep learning based methods are proposed for capturing discriminative and generalized visual appearance~\cite{yu2020searching,yu2020face,Atoum2018Face,cai2020drl} and physiological rPPG~\cite{Liu2018Learning,yu2021transrppg} features.

Besides the physical-world face spoofing attacks, with unrestricted access to the rapid proliferation of digital face images on social media platforms, launching forgery attacks against face recognition systems has become even more accessible. Given the growing dissemination of `fake news' and `deepfakes'~\cite{tolosana2020deepfakes}, both research community and social media platforms are pushing towards generalizable detection techniques against sophisticated forgery attacks. 
Early works are based on traditional digital forensics features~\cite{buchana2016simultaneous,fridrich2012rich} while later ones rely heavily on deep learning based image and video classification (real vs. fake)~\cite{li2020face,qian2020thinking} using the visual appearance deepfake inputs. However, deepfake images and videos are  continuously evolving and become more and more visually realistic due to the more mature and powerful deep image generation methods~\cite{creswell2018generative}. To tackle the detection issue under weak and unperceived appearance artifacts, physiological rPPG based methods built on traditional time-frequency statistics~\cite{ciftci2020fakecatcher,hernandez2022deepfakes} and deep rhythm features~\cite{qi2020deeprhythm,sun2022faketransformer,ciftci2020hearts} are proposed.

\begin{figure*}
\centering
\includegraphics[scale=0.28]{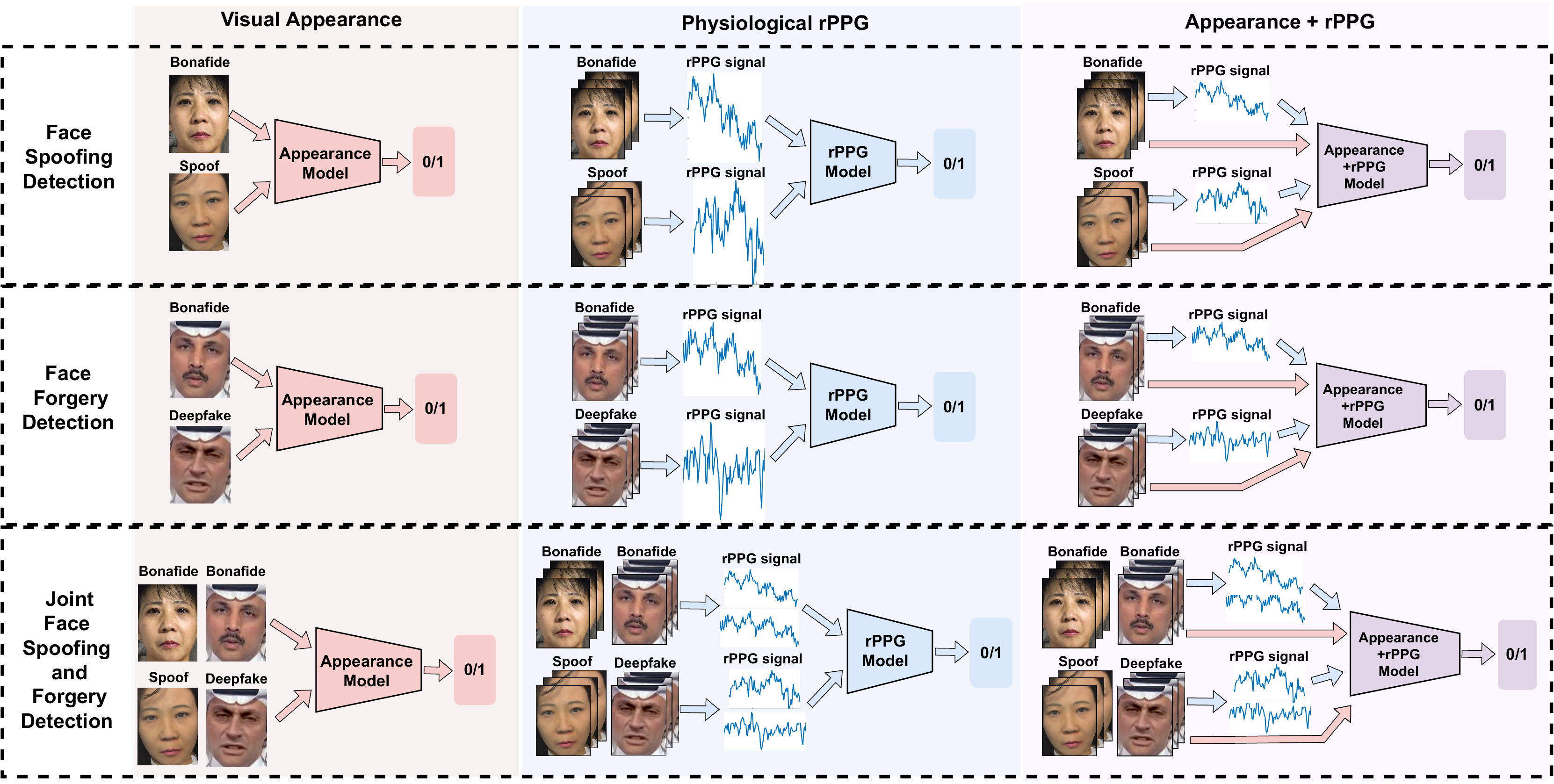}
\vspace{-0.4em}
  \caption{\small{
  Visualization of the modality (visual appearance and physiological rPPG) and task (face spoofing and forgery detection) matrix. Columns from left to right: 1) appearance models with RGB face inputs; 2) rPPG models with facial rPPG signals inputs; and 3) appearance+rPPG models with both RGB face and facial rPPG inputs. Rows from top to bottom: 1) separate face spoofing detection with bonafide/spoof for training; 2) separate face forgery detection with bonafide/deepfakes for training; and 3) joint face spoofing and forgery detection with bonafide/spoof/deepfakes for training. Best views in color.
  }
  }
\vspace{-1.2em}
\label{fig:figure1}
\end{figure*}

Despite excellent achievements in visual appearance or physiological rPPG based unimodal face attack detection, single-modal models cannot generalize well to various types of attacks. As for the FAS and face forgery detection tasks, facial appearance features cannot generalize well under unseen scenarios of artifact disappearing while hidden rPPG clues are less reliable when large head motions are included and heavy compression is employed. Compared to single-modal ones, multi-modal models on both appearance and rPPG cues attract less attention due to the lack of mature fusion strategies. In this case, detection based on either visual appearance or physiological rPPG cues might be sub-optimal. To the best of our knowledge, there is only one work~\cite{lin2019face} to explore the feasibility of combing appearance and rPPG predictions via score-level fusion for generalized FAS. However, it is still unknown: 1) how multi-modal models perform on the face forgery detection task; and 2) whether sophisticated feature-level fusions help boost the performance.

In terms of the relationship between FAS and face forgery detection tasks, bonafides inherently share the similar clean appearance and periodic rPPG cues while spoofs/deepfakes have different kinds of artifacts and noisy rhythms. In other words, these two tasks are highly correlated in both visual~\cite{deb2021unified} and physiological modalities, inspiring us to explore whether multi-task learning benefits the common and intrinsic feature representation in both unimodal (appearance or rPPG) and multi-modal (appearance+rPPG) settings.

Motivated by the above discussions, we propose a joint face spoofing and forgery detection benchmark using both visual appearance and physiological rPPG cues. The frameworks of face attack detection based on the inputs of facial appearance, rPPG, and appearance+rPPG are shown in the first (red), second (blue), and third (purple) columns of Fig.~\ref{fig:figure1}, respectively. From the perspective of separate/joint attack detection, tasks shown in the first and second rows of Fig.~\ref{fig:figure1} represent the separate FAS and face forgery detection, respectively. And the joint face spoofing and forgery detection task is illustrated in the last row of Fig.~\ref{fig:figure1}, where the bonafide, spoofing, and deepfake samples are all included for training. In this paper, we generally define the bonafide as live samples from FAS task and genuine ones from face forgery detection task, while the spoofing and deepfakes denote physical presentation attacks in FAS task and digital manipulation attacks in face forgery detection task, respectively.  

Compared with well-studied appearance based FAS and face forgery detection methods, the ones based on rPPG or appearance+rPPG cues are still immature. On one hand, the periodicity discrimination of the time-domain rPPG signals is limited due to external interference (e.g., video quality and head movement). In other words, under serious interference, the rPPG curves from the bonafide might still hold noisy temporal shapes like those from attacks. On the other hand, due to the modality bias (performance and distribution) between the appearance and rPPG features, the direct fusion is unstable and might has negative effects. To tackle the first issue, an extra network branch with continuous wavelet transformed time-frequency representation is introduced, which is able to directly provide useful dynamic frequency variation cues to enhance the periodicity discrimination. Furthermore, to alleviate the modality bias and improve the fusion efficacy, we propose to adopt a weighted batch and layer normalization for separate modality features refinement before fusion. 

In terms of multi-task learning for joint face spoofing and forgery detection, we conduct elaborate studies on prevalent deep models and learning configurations, and find some valuable conclusions: 1) joint face spoofing and forgery detection can benefit the generalization capacities of \textit{most prevalent deep models} and \textit{both modalities (i.e., appearance and rPPG)} on cross-dataset testings for both tasks; 2) task sampling strategy influences the multi-task learning a lot, and the simplest \textit{random sampling} between two tasks surprisingly works best; 3) joint training with \textit{both bonafide and attack samples} performs the best while with only extra attack samples from the other task might degrade the performance; and 4) backbone sharing in \textit{low and mid levels} benefits the multi-task learning.

To sum up, our contributions include:

\begin{itemize}
    \item We establish the first joint face spoofing and forgery detection benchmark where both visual appearance and physiological rPPG cues are fully exploited.  
    
    \item We design a two-branch physiological network using both facial spatio-temporal rPPG signal map and its continuous wavelet transformation counterpart as inputs to enhance the rPPG periodicity discrimination.
    
    \item We conduct a weighted batch and layer normalization for both appearance and rPPG features before multi-modal fusion to mitigate the modality bias and improve the fusion efficacy.
    
    \item We investigate prevalent deep models, fusion strategies and multi-task learning configurations for joint face spoofing and forgery detection, and find that the generalization capacities of both unimodal (appearance or rPPG) and multi-modal (appearance+rPPG) models can be obviously improved via joint training on both tasks.

\end{itemize}

In the rest of the paper, Section~\ref{sec:relatedwork} provides the related work. Section~\ref{sec:method} formulates the joint FAS and face forgery detection. Then, it introduces the two-branch physiological framework with CWT map and the fusion strategy with weighted batch and layer normalizations. Section~\ref{sec:benchmark} provides the details of the benchmark establishment. Section~\ref{sec:experiment} provides rigorous ablation studies and evaluates the performance of mainstream models on the proposed benchmark. Finally, a conclusion with future directions is given in Section~\ref{sec:conclusion}.


\vspace{-1.0em}
\section{Related Work}
\label{sec:relatedwork}
\vspace{-0.3em}


\subsection{Face Anti-Spoofing}
 \textbf{Visual appearance based FAS.}\quad  
Traditional face anti-spoofing (FAS) methods usually extract hand-crafted texture descriptors, e.g., LBP~\cite{boulkenafet2015face} and HOG~\cite{Komulainen2014Context}, from the facial images to capture the spoofing patterns. Some deep learning based methods treat the visual appearance based FAS problem as a binary classification task~\cite{Li2017An}, and utilize binary cross-entropy loss to optimize the model. Motivated by physical discrepancy between bonafide and spoof faces, dense pixel-wise supervisions~\cite{yu2021revisiting}, e.g., pseudo depth map~\cite{Liu2018Learning,yu2020searching,wang2020deep,yu2021dual}, reflection map~\cite{zhang2020celeba,yu2020face}, texture map~\cite{zhang2020face} and binary map~\cite{george2019deep,liu2019deep} are introduced for learning intrinsic features. 
FAS can be also formulated as domain adaptation or generalization problems in~\cite{li2018domain, jia2020single, wang2020cross}. Disentangled learning~\cite{zhang2020face,liu2020disentangling}, adversarial learning~\cite{shao2019multi}, and meta learning~\cite{shao2019regularized,cai2022learning} are usually adopted to improve the model's generalization capacity on unseen scenarios. Besides, several anomaly detection~\cite{li2020unseen,nikisins2018effectiveness}, zero-shot~\cite{liu2019deep,qin2021meta} and continuous learning~\cite{rostami2021detection} approaches are proposed for appearance based unknown PA detection.

 \textbf{Physiological rPPG based FAS.}\quad 
As facial-video based remote rPPG measurement techniques achieve great developments~\cite{yu2022physformer,yu2019remote}, many researchers utilize rPPG to detect face spoofing attacks. Li et al.~\cite{li2016generalized} firstly consider the difference of pulse between bonafide faces and printed faces for spoofing attack detection. Several works~\cite{nowara2017ppgsecure,liu20163d,liu2021multi} compare facial rPPG signals and background noises to decide whether the faces contain liveness clues or not. Moreover, combining with deep learning methods, Yu et al~\cite{yu2021transrppg} propose an rPPG transformer framework (TransRPPG) to extract global periodicity for FAS. Liu et al.~\cite{liu2019deep} utilize the generated pseudo rPPG frequency distributions as auxiliary supervision signals to benefit discriminative dynamic clues representation. Instead of using only the rPPG modality, Lin et al.~\cite{lin2019face} explore the feasibility of combing appearance and rPPG predictions via score-level fusion for generalized FAS.

\vspace{-1.0em}
\subsection{Face Forgery Detection}
 \textbf{Visual appearance based face forgery detection.}\quad  
Many efforts have been made to improve the performance of face forgery detection~\cite{tolosana2020deepfakes,yang2021mtd,li2020face}. Early works~\cite{nguyen2019capsule,rossler2019faceforensics} use generic deep models (e.g., Xception~\cite{chollet2017xception}) to extract features from cropped face images and perform binary classification, which is weak in capturing fine-grained artifacts between bonafide and forged images. In view of that forged faces become more visually realistic, more reliable forgery patterns such as noise statistics, local textures, and frequency information are proposed. For example, Zhou et al.~\cite{zhou2017two} design a two-stream network to combine visual appearance and local noise patterns for forgery detection. Zhao et al.~\cite{zhao2021multi} propose a multi-attentional face forgery detector to efficiently aggregate the texture and high-level semantic features from multiple local regions. Several other works~\cite{qian2020thinking,li2021frequency,liu2021spatial} also consider the frequency details into account and propose frequency-aware models to distinguish bonafide faces from forged faces. 

 \textbf{Physiological rPPG based face forgery detection.}\quad 
Similar to the rPPG based FAS methods, rPPG based face forgery detection~\cite{hernandez2022deepfakes,hernandez2020deepfakeson} mainly relies on the judgment about whether the heart rhythm patterns contain solid live/periodic patterns since the heart rhythms extracted from forged face videos are usually diminished by deepfake methods, and are with less periodic cues compared with those from bonafide faces. Ciftci et al~\cite{ciftci2020fakecatcher} extract rich expert-designed statistical rPPG features from both the time domain and power spectra domain to describe spatial coherence and temporal consistency for deepfake detection. Subsequently, several transformed rPPG maps such as PPG cell~\cite{ciftci2020hearts}, motion-magnified spatio-temporal map~\cite{qi2020deeprhythm}, and Gaussian spatial decomposition map~\cite{sun2022faketransformer} are proposed, which are cascaded with deep CNN or transformer models for bonafide/forgery binary classification.

\vspace{-1.2em}
\subsection{Multi-Task Learning for Face Attack Detection}
\vspace{-0.1em}

Multi-task learning aims at developing models that can address multiple tasks via sharing partial parameters and computation~\cite{caruana1997multitask}. Numerous works have observed that although multi-task models are more versatile, their accuracies are easily lower than single-task models, especially when simultaneously performing unrelated tasks~\cite{kokkinos2017ubernet,zamir2018taskonomy}. Moreover, jointly training models to simultaneously perform multiple tasks has typically required careful calibration of the individual tasks, to ensure that none of the task-specific losses dominates another. Methods to mitigate this issue include gradient-normalization~\cite{chen2018gradnorm} and adaptive loss weights~\cite{kendall2018multi}. 

In the field of FAS, few works~\cite{wu2020uncertainty,yu2020face} are proposed to mine task-shared intrinsic features (i.e., material property of generic objects and spoof instruments) via multi-task learning between visual material recognition and FAS tasks. However, there are still huge task-aware biases (e.g., appearance and physiological cues) between generic objects and human faces. In terms of unified digital and physical face attack detection, one recent work~\cite{deb2021unified} proposes to adaptively learn joint appearance representations for coherent attacks, while uncorrelated attack types are learned separately. All previous works focus on the visual appearance modality but ignore the rPPG based multi-task learning. To fill in the blank, we establish the first joint face spoofing and forgery detection benchmark using both visual appearance and physiological rPPG cues to explore whether more common and intrinsic features can be learned on such two highly-correlated tasks. 

\vspace{-0.8em}
\section{Methodology}
\label{sec:method}

In this section, we first introduce the formulation of joint face spoofing and forgery detection in Section~\ref{sec:joint}. To enhance the rPPG periodicity discrimination, we then propose a two-branch physiological network in Section~\ref{sec:rppg}. Finally, to improve the fusion efficacy between appearance and rPPG features, we present a simple yet effective weighted batch and layer normalization strategy in Section~\ref{sec:normalization}.

\textbf{Preliminary.}\quad   Followed by most previous works, we treat face spoofing and forgery detection tasks as a binary (bonafide/attack) classification problem to train models with binary cross-entropy (BCE) loss~\cite{ming2022vitranspad,rossler2019faceforensics} or binary pixel-wise supervision~\cite{george2019deep,yu2021revisiting}. Given an extracted face $X_{App}$ or rPPG signal map $X_{rPPG}$ input, deep appearance features $F_{App}$ or rPPG features $F_{rPPG}$ can be represented via forwarding the corresponding unimodal models $\Phi_{App}$ and $\Phi_{rPPG}$, respectively. Then the cascaded classification heads make the binary predictions $Y_{Pre}$, which are supervised by the BCE loss
\begin{equation} 
\mathcal{L}_{BCE}=-(Y_{GT}log(Y_{Pre})+(1-Y_{GT})log(1-Y_{Pre})),
\end{equation}
where $Y_{GT}$ is the ground truth ($Y_{GT}=0$ for attack and $Y_{GT}=1$ for bonafide. Similarly, for the models~\cite{george2019deep} supervised with binary pixel-wise supervision, the BCE loss is accumulated on all spatial dimensions.

\vspace{-0.8em}
\subsection{Joint Face Spoofing and Forgery Detection}
\label{sec:joint}

In general, separate models (i.e., $\Phi^{Spoof}$ and $\Phi^{Forgery}$) are trained for face spoofing and forgery detection. Such separate training for each task usually neglects the high correlation between these two tasks. On one hand, bonafides reasonably share the similar clean appearance and periodic rPPG cues. Learning upon the merged bonafide samples from both two tasks is able to provide more valid samples to alleviate overfitting for bonafide representation. On the other hand, spoofs and deepfakes have different kinds of artifacts and noisy rhythms, which enhances the diversity of the attack data. Thus, learning from the merged diverse attack samples might improve the models' generalization. Instead of training within each standalone task, we propose to train a unified model $\Phi^{Joint}$ in a multi-task learning fashion.

\begin{figure}[t]
\centering
\includegraphics[scale=0.65]{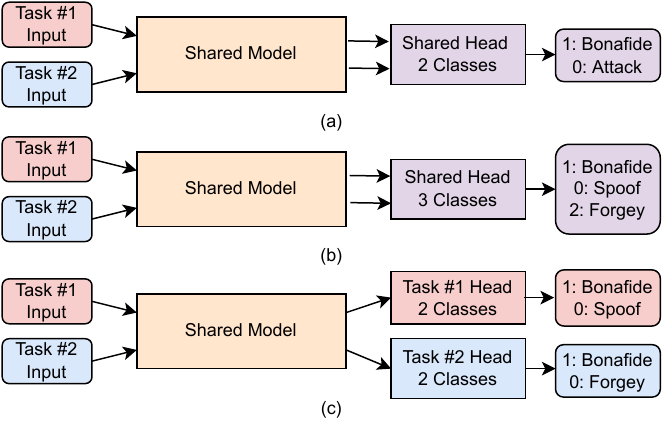}
\vspace{-1.0em}
 \caption{\small{
  Different head settings for multi-task learning with (a) a shared single head for binary classification; (b) a shared single head for 3-class classification; and (c) separate heads for binary classification. Task \#1 and \#2 indicate face spoofing and forgery detection, respectively. }
  }
  \vspace{-1.5em}
\label{fig:head}
\end{figure}

\textbf{Joint Training Architecture.}\quad  Here we consider the simplest setting (i.e., all backbones and classification heads are shared across two tasks) first. In this case, there are no extra learnable task-specific parameters thus it is as efficient as single-task models for inference. Specifically, the shared heads with 2 classes (bonafide/attack) and 3 classes (bonafide/spoof/forgery) are shown in Fig.~\ref{fig:head}(a) and (b), respectively. Compared with the former one supervised via the BCE loss and making a binary prediction, the latter one further distinguishes the attack class into two more fine-grained task-specific classes (i.e., spoof and forgery), and is supervised via the 3-class cross-entropy loss. Besides the shared head setting, there is another favorite setting to cascade task-specific heads behind the shared model (see Fig.~\ref{fig:head}(c)) to flexibly project the shared embedding features into a separate decision space. Thus, the classification heads are the only task-specific parameters. The study about the head setting can be found in Table~\ref{tab:heads}.

To increase model representation capacity (with $N$ layers) when jointly training on two tasks simultaneously, we can optionally include $N_{shared}>0$ task-shared layers in front while cascading $N_{specific}=N-N_{shared}$ task-specific layers. In this case, lower-level detailed features (e.g., edge, contour, color and quality) are shared across two tasks to alleviate overfitting while task-specific higher-level semantic features (e.g., moire pattern and spatial inconsistency) are learned separately to enhance the discrimination ability. We study the impacts of $N_{shared}$ for both appearance and rPPG models in Fig.~\ref{fig:Ablation12}(a)-(d).

\textbf{Joint Task Sampling Strategy.}\quad   We optimize the unified model parameters $\Phi^{Joint}$ simultaneously across two tasks  with stochastic gradient descent (SGD). As a result, there are different sampling strategies for task-related training batch construction. During joint training, for each SGD step, we sample a minibatch from the merged two-task data $D_{Spoof}\cup D_{Forgery}$, and then evaluate a gradient for updating $\Phi^{Joint}$. An important consideration is the proportion of per-task data in each minibatch and task sampling order. We describe several task sampling schedules below. $U_{Spoof}$ and $U_{Forgery}$ are denoted as the number of SGD steps in each epoch for single-task face spoofing and forgery detection, respectively.

\begin{itemize}
    \item Random. In this schedule, each batch is randomly sampled from the merged two-task data $D_{Spoof}\cup D_{Forgery}$, thus the proportion of per-task data is stochastic. 
    
    \item Simultaneous. Here we sample the minibatch always encompassing data from both tasks simultaneously. Specifically, a half of the minibatch is assembled with data from $D_{Spoof}$ while the other half is from $D_{Forgery}$.
    
    \item Alternating. This deterministic schedule alternates between $D_{Spoof}$ and $D_{Forgery}$ in a fixed, repeating order. Concretely, we perform a single SGD step for each task in sequence before repeating the same order.
    
    \item Task-by-task. In this schedule, the first $U_{Spoof}$ SGD steps are performed with FAS task. Then the next $U_{Forgery}$ steps with face forgery detection task are performed and vice versa.
\end{itemize}

Random sampling is adopted as the default minibatch construction strategy due to its simplicity and satisfactory performance. Detailed studies can be found in Table~\ref{tab:sampling}.

\vspace{-0.8em}
\subsection{Two-branch Physiological Network }
\label{sec:rppg}

\begin{figure}[t]
\centering
\includegraphics[scale=0.33]{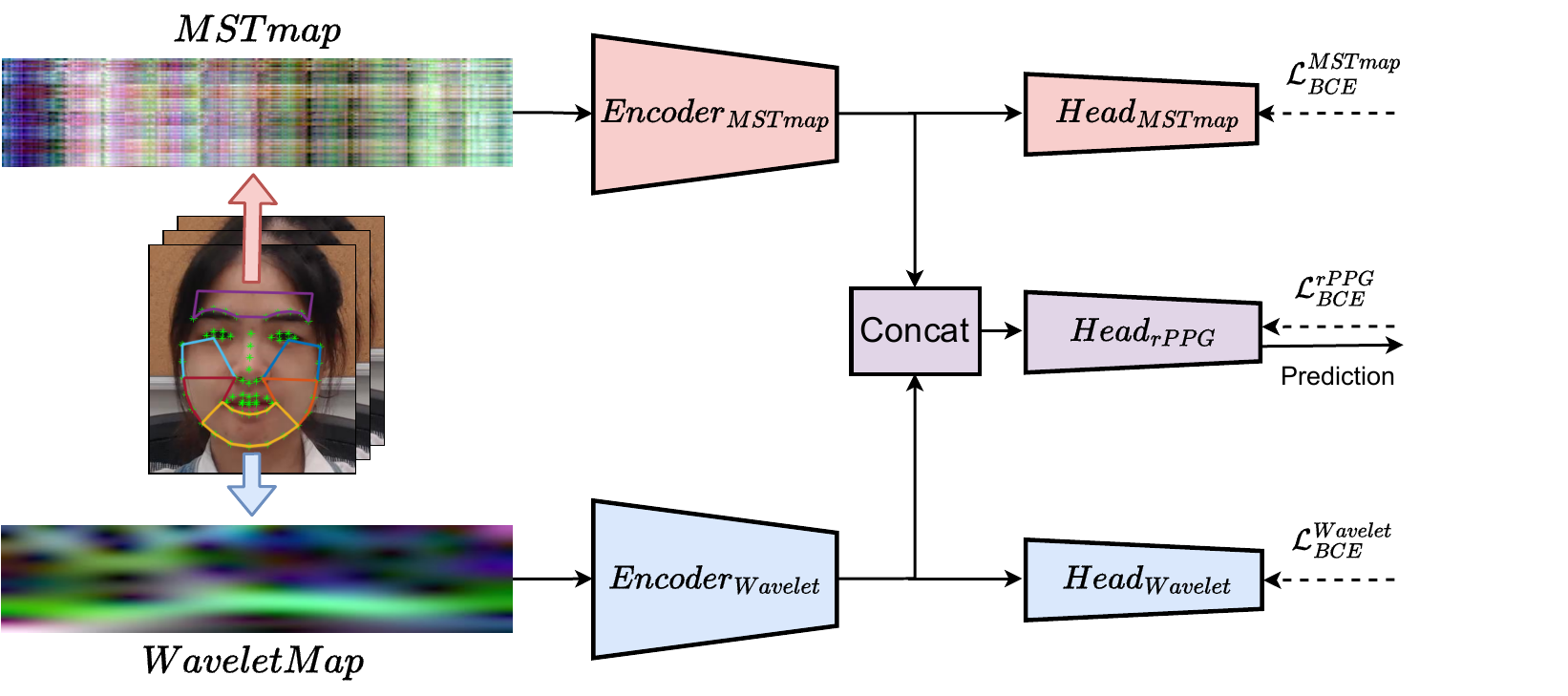}
\vspace{-1.6em}
 \caption{\small{
  The framework of the two-branch physiological network. The facial time-domain based MSTmap and wavelet transformed time-frequency representation WaveletMap are used as two-branch inputs. }
  }
  \vspace{-1.6em}
\label{fig:rPPG}
\end{figure}

The amplitude of optical absorption variation of face skin is very small, thus it is important to design a good representation to highlight the physiological information in face videos for robust face spoofing and forgery detection. To this end, we utilize both the multi-scale spatial-temporal map (MSTmap)~\cite{niu2020video} and the continuous wavelet transformation (CWT) based time-frequency map as two-branch inputs for the physiological network. The overall framework is illustrated in Fig.~\ref{fig:rPPG}.

\textbf{MSTmap and WaveletMap.}\quad   The MSTmap considers both the local and the global physiological information to represent the facial skin color variations due to heartbeats. For the $t$-th frame of a face video, we first get a set of $K$ informative regions of face $R_{t} = \left\{ R_{1t},R_{2t},...,R_{Kt} \right\}$. Then, we calculate the average pixel values of each RGB color channel for all the non-empty subsets of $R_{t}$, which are $2K−1$ combinations of the elements in $R_{t}$. Therefore, given a facial video with $T$ frames, an MSTmap with the size of $(2K-1) \times T \times 3$ is obtained, where the three dimensions mean the numbers of total regions, frames, and color channels.

Besides the representation in the time domain, we also exploit  
the wavelet representation of rPPG signals to form a time-frequency view, which contains straightforward and rich periodicity patterns. Specifically, the CWT of the global rPPG signal $S(t)$ (i.e., averaged from all $K$ facial regions) corresponds to a time-frequency representation computed from a prototype function, i.e., mother wavelet, which can be formulated as
\vspace{-0.5em}
\begin{equation}
\begin{split}
&CWT_{S}^{\psi }\left ( \tau ,d \right )=\int_{-\infty }^{\infty }S(t)\psi_{\tau ,d}(t)dt,\\
&\psi_{\tau ,d}(t)=\frac{1}{\sqrt{\left|d \right|}}\psi(\frac{t-\tau}{d}),
\end{split}
\label{eq:wavelet1}
\end{equation}
where $\psi_{\tau ,d}$ corresponds to the mother wavelet dilated by $d$ and translated by $\tau$. Unlike the Fourier transform, the wavelet transform can detect abrupt changes in frequency using a family of wavelets $\psi_{\tau ,d}$ computed from the mother wavelet $\psi$. We use the Matlab implementation `cwtfilterbank' with Morse wavelet and 48 voices per octave.

\begin{figure*}[t]
\centering
\includegraphics[scale=0.32]{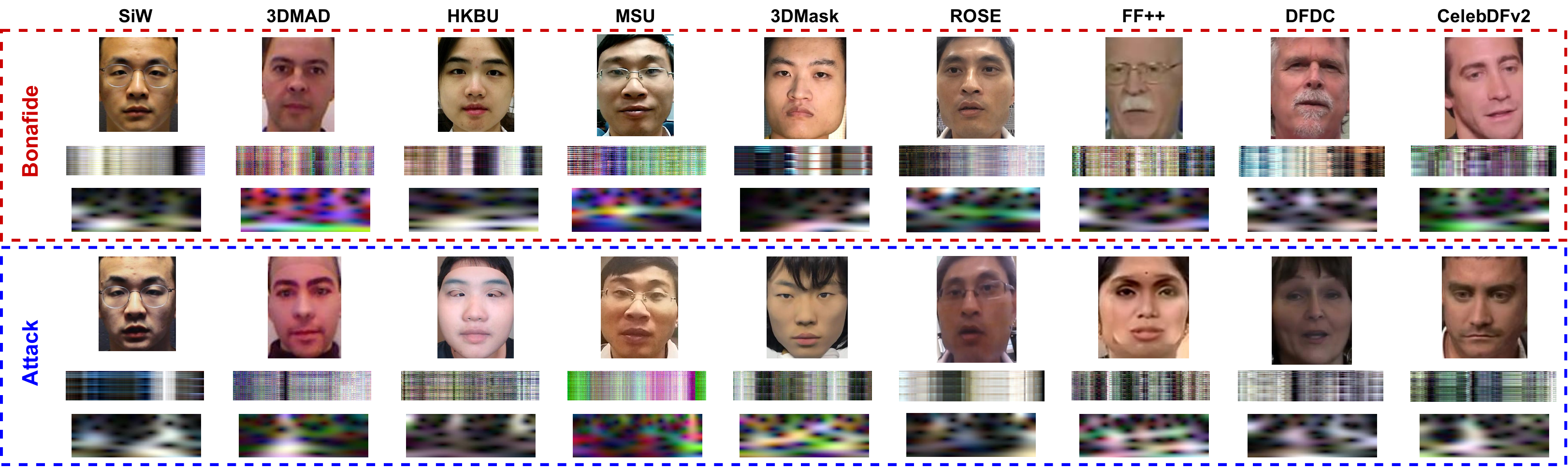}
\vspace{-0.5em}
 \caption{\small{Representative visual (RGB faces in the first and fourth rows) samples as well as their rPPG maps (MSTmap~\cite{niu2020video} in the second and fifth rows while wavelet maps in the third and sixth rows) on nine benchmark datasets.}
  }
 \vspace{-0.8em}
\label{fig:protocol}
\end{figure*}

\textbf{Two-branch Network.}\quad  Unlike the previous works~\cite{yu2021transrppg,qi2020deeprhythm} only considering the time-domain input, the proposed two-branch network adopts the extracted facial MSTmap and WaveletMap as inputs. As shown in Fig.~\ref{fig:rPPG}, the MSTmap and WaveletMap first forward through the same but parameter-unshared encoders to obtain the time-domain and time-frequency-domain deep features, respectively. Then these two kinds of features are concatenated to form the generalized rPPG features. Finally, a binary classification head supervised by the BCE loss $\mathcal{L}^{rPPG}_{BCE}$ is cascaded for prediction. To benefit the training stability, we also add auxiliary binary classification heads for each separate branch, which are supervised by the BCE losses $\mathcal{L}^{MSTmap}_{BCE}$ and $\mathcal{L}^{Wavelet}_{BCE}$, respectively. The overall loss is denoted as $\mathcal{L}^{rPPG}_{overall}=\mathcal{L}^{rPPG}_{BCE}+\alpha\cdot(\mathcal{L}^{MSTmap}_{BCE}+\mathcal{L}^{Wavelet}_{BCE})$, where hyperparameter $\alpha=0.5$. By leveraging both the fine-grained time-domain color change rhythms and the semantic time-frequency distribution, the two-branch physiological network is able to capture discriminative and generalized hidden periodicity cues.  


\vspace{-1.0em}
\subsection{Weighted Normalization for Appearance and rPPG Fusion}
\label{sec:normalization}

As for the appearance+rPPG multi-modal models, the naive framework is to fuse the appearance features $F_{App}$ and rPPG features $F_{rPPG}$ first, and then a prediction head with the BCE loss $\mathcal{L}^{fuse}_{BCE}$ is cascaded behind the fused multi-modal features $F_{fuse}$. The overall loss for the multi-modal model supervision can be formulated as $\mathcal{L}^{fuse}_{overall}=\mathcal{L}^{fuse}_{BCE}+\beta\cdot( \mathcal{L}^{App}_{BCE}+\mathcal{L}^{rPPG}_{overall})$, where hyperparameter $\beta=0.5$.

Due to the modality discrepancy between visual appearance and physiological rPPG, their corresponding hidden features ($F_{App}$ and $F_{rPPG}$) usually differ significantly, making the fusion challenging (e.g., the direct feature summation drops the performance compared with unimodal models). One straightforward solution is to concatenate these two features in the channel domain first~\cite{yu2020multi}, and then aggregate the multi-modal heterogeneous features with a lightweight fusion operator (e.g., linear fully connection layer). The direct concatenation fusion can be formulated as    
\vspace{-0.1em}
\begin{equation} 
F_{fuse} = \mathrm{ReLU(Linear(Concat}(F_{App},F_{rPPG}))).
\end{equation}

However, feature misalignment among modalities leads to inefficient feature concatenation fusion. To alleviate this issue, we propose to utilize weighted batch normalization (BN)~\cite{ioffe2015batch} and layer normalization (LN)~\cite{ba2016layer} for each independent modality branch first to map $F_{App}$ and $F_{rPPG}$ into the aligned subspace. Then the refined features ($F^{Norm}_{App}$ and $F^{Norm}_{rPPG}$ are concatenated and aggregated. The weighted normalization fusion can be formulated as    
\begin{equation}
\begin{split}
&F^{Norm}_{App} = \theta \cdot LN(F_{App}) + (1-\theta) \cdot BN(F_{App}),\\
&F^{Norm}_{rPPG} = \theta \cdot LN(F_{rPPG}) + (1-\theta) \cdot BN(F_{rPPG}),\\
&F_{fuse} = \mathrm{ReLU(Linear(Concat}(F^{Norm}_{App},F^{Norm}_{rPPG}))),
\end{split}
\label{eq:normalization}
\end{equation}
where hyperparameter $\theta \in [0,1]$ tradeoffs the contribution between BN and LN, which is studied in Fig.~\ref{fig:Ablation12}(e)(f). As the embedding features $F_{App}$ and $F_{rPPG}$ are already projected among all spatial dimensions (e.g., average pooling in ResNet while global class token in ViT), the feature statistics (i.e., mean and variance) from BN and LN are only calculated for each individual channel across all batch samples and each individual sample across all channels, respectively. The merits of the weighted normalization fusion are two-fold: 1) towards efficient fusion based on the refined modality features via leveraging informative statistics from two views (batch and channel); and 2) enhancing the stability and benefiting the convergence for multi-modal model training.

To verify the efficacy of the weighted batch and layer normalization for appearance and rPPG feature fusion, two other fusion strategies (i.e., squeeze-and-excitation (SE) fusion~\cite{hu2018squeeze,casiasurf} and cross-attention fusion~\cite{yu2022flexible}) are compared (see Table~\ref{tab:fusion}). Please note that in this paper, we focus on feature-level fusion strategies but there are also e.g., input-level fusion and decision-level fusion for multi-modal scenarios, which might be explored in future work.

\section{Joint Face Spoofing and Forgery Detection Benchmark}
\label{sec:benchmark}

In this section, we introduce the joint face spoofing and forgery detection benchmark in terms of datasets, protocols, and evaluation metrics. A statistical description of the datasets is shown in Table~\ref{tab:protocol}.  

\vspace{0.1em}
  \textbf{Datasets.} \quad  In consideration of the efficacy of the long-term physiological rPPG cues~\cite{liu20163d,yu2021transrppg}, we select nine datasets (i.e., SiW~\cite{Liu2018Learning}, 3DMAD~\cite{erdogmus2014spoofing}, HKBU-MarsV2 (HKBU)~\cite{liu20163d}, MSU-MFSD (MSU)~\cite{wen2015face}, 3DMask~\cite{yu2020fas2} and ROSE-Youtu (ROSE)~\cite{li2018unsupervised} for FAS while FaceForensics++ (FF++)~\cite{rossler2019faceforensics}, DFDC~\cite{dolhansky2019deepfake} and  CelebDFv2~\cite{li2020celeb} for face forgery detection) containing long videos (mostly $\textgreater$5 seconds) for the joint face spoofing and forgery detection benchmark, which is suitable for investigations of both visual appearance and rPPG modalities. We conduct both intra-domain and cross-domain experiments for benchmark evaluation. SiW, 3DMAD, HKBU, and FF++ are used for intra-domain evaluation. The original train/test splits in~\cite{Liu2018Learning,rossler2019faceforensics} are strictly followed for SiW and FF++. In consideration of no original splits for 3DMAD and HKBU datasets, we simply split the samples from the first 11 and 6 subjects for training while the remaining for testing. Besides, the whole MSU, 3DMask , ROSE, DFDC, and CelebDFv2 datasets are used for cross-domain evaluation. Representative visual samples as well as their rPPG maps are illustrated in Fig.~\ref{fig:protocol}.

\begin{table}[t]
\centering
\caption{The datasets and their corresponding properties used in the joint face spoofing and forgery detection benchmark. `Sub.', `Bona.' and `Att.' are short for subjects, bonafide and attack, respectively.}
\vspace{-0.5em}
\resizebox{0.49\textwidth}{!}{
\begin{tabular}{|ccccc|}
\hline
\multicolumn{1}{|c|}{Dataset}   & \multicolumn{1}{c|}{\begin{tabular}[c]{@{}c@{}}Train Videos \\ (Bona./Att.)\end{tabular}} & \multicolumn{1}{c|}{\begin{tabular}[c]{@{}c@{}}Test Videos\\ (Bona./Att.)\end{tabular}} & \multicolumn{1}{c|}{\# Sub.} & Attack Types                                                                                  \\ \hline
\multicolumn{5}{|c|}{\cellcolor[HTML]{EFEFEF}Face Spoofing Detection}                                                                                                                                                                                                                                                                                \\ \hline
\multicolumn{1}{|c|}{SiW~\cite{Liu2018Learning}}       & \multicolumn{1}{c|}{714/1711}                                                             & \multicolumn{1}{c|}{599/1462}                                                           & \multicolumn{1}{c|}{165}     & 2 (Print, Replay)                                                                             \\ \hline
\multicolumn{1}{|c|}{3DMAD~\cite{erdogmus2014spoofing}}     & \multicolumn{1}{c|}{110/55}                                                               & \multicolumn{1}{c|}{60/30}                                                              & \multicolumn{1}{c|}{17}      & 1 (Mask)                                                                                      \\ \hline
\multicolumn{1}{|c|}{HKBU~\cite{liu20163d}}      & \multicolumn{1}{c|}{336/336}                                                              & \multicolumn{1}{c|}{252/252}                                                            & \multicolumn{1}{c|}{12}      & 1 (Mask)                                                                                      \\ \hline
\multicolumn{1}{|c|}{MSU~\cite{wen2015face}}       & \multicolumn{1}{c|}{-}                                                                    & \multicolumn{1}{c|}{70 / 210}                                                           & \multicolumn{1}{c|}{35}      & 2 (Print, Replay)                                                                             \\ \hline
\multicolumn{1}{|c|}{3DMask~\cite{yu2020fas2}}    & \multicolumn{1}{c|}{-}                                                                    & \multicolumn{1}{c|}{288 / 864}                                                          & \multicolumn{1}{c|}{48}      & 1 (Mask)                                                                             \\ \hline
\multicolumn{1}{|c|}{ROSE~\cite{li2018unsupervised}}      & \multicolumn{1}{c|}{-}                                                                    & \multicolumn{1}{c|}{500 / 2850}                                                         & \multicolumn{1}{c|}{20}      & 3 (Print, Replay, Paper Mask)                                                                       \\ \hline
\multicolumn{5}{|c|}{\cellcolor[HTML]{EFEFEF}Face Forgery Detection}                                                                                                                                                                                                                                                                                 \\ \hline
\multicolumn{1}{|c|}{FF++~\cite{rossler2019faceforensics}}      & \multicolumn{1}{c|}{720/2880}                                                             & \multicolumn{1}{c|}{140/560}                                                            & \multicolumn{1}{c|}{-}       & \begin{tabular}[c]{@{}c@{}}4 (Face2Face, FaceSwap, \\ Deepfakes, NeuralTextures)\end{tabular} \\ \hline
\multicolumn{1}{|c|}{DFDC~\cite{dolhansky2019deepfake}}      & \multicolumn{1}{c|}{-}                                                                    & \multicolumn{1}{c|}{1131 / 4113}                                                        & \multicolumn{1}{c|}{-}       & 2 (Deepfakes)                                                                                 \\ \hline
\multicolumn{1}{|c|}{CelebDFv2~\cite{li2020celeb}} & \multicolumn{1}{c|}{-}                                                                    & \multicolumn{1}{c|}{590 / 5639}                                                         & \multicolumn{1}{c|}{59}      & 1 (Deepfakes)                                                                                 \\ \hline
\end{tabular}}
\label{tab:protocol}
\vspace{-1.2em}
\end{table}

\begin{table*}[]
\centering
\caption{Results of appearance-based unimodal models on the joint face spoofing and forgery detection benchmark. `Sep.' is short for separate training. The higher AUC(\%) $\uparrow$ and TPR(\%) $\uparrow$ while the lower EER(\%) $\downarrow$, the better performance. Best cross-testing results are marked in \textbf{bold}. } \label{tab:appearance}
\vspace{-0.5em}
\resizebox{0.88\textwidth}{!}{
\begin{tabular}{|c|c|c|cccccc|ccc|}
\hline
\multirow{3}{*}{Metrics}                                                                    & \multirow{3}{*}{\begin{tabular}[c]{@{}c@{}}Sep./\\ Joint\end{tabular}} & \multirow{3}{*}{Method}             & \multicolumn{6}{c|}{Face Spoofing Detection}                                    & \multicolumn{3}{c|}{Face Forgery Detection}                             \\ \cline{4-12} 
                                                                                            &                                                                        &                                     & \multicolumn{3}{c|}{Intra-testing}         & \multicolumn{3}{c|}{Cross-testing} & \multicolumn{1}{c|}{Intra-testing} & \multicolumn{2}{c|}{Cross-testing} \\ \cline{4-12} 
                                                                                            &                                                                        &                                     & SiW   & 3DMAD & \multicolumn{1}{c|}{HKBU}  & MSU       & 3DMask     & ROSE      & \multicolumn{1}{c|}{FF++}          & DFDC          & Celeb-DFv2         \\ \hline
\multirow{15}{*}{AUC(\%)}                                                                   & \multirow{7}{*}{Sep.}                                                  & MesoNet                             & 95.73 & 100   & \multicolumn{1}{c|}{68.87} & 85.33     & 44.54      & 68.05     & \multicolumn{1}{c|}{65.54}         & 54.72         & 65.1               \\ 
                                                                                            &                                   & Xception                            & 99.88 & 100   & \multicolumn{1}{c|}{94.47} & 90.62     & 63.88      & 90.15     & \multicolumn{1}{c|}{98.21}         & 64.86         & 84.27              \\                                    
                                                                                            &                                                                        & MultiAtten                          & 99.56 & 100   & \multicolumn{1}{c|}{68.68} & 89.21     & 36.88      & 72.66     & \multicolumn{1}{c|}{70.24}         & 63.37         & 64.34              \\
                                                                                            &                                                                        & CDCN++                              & 100   & 99.56 & \multicolumn{1}{c|}{94.51} & 82.65     & 47.66      & 76.76     & \multicolumn{1}{c|}{97.53}         & 56.65         & 78.34              \\
                                                                                            &                                                                        & DeepPixel                           & 100   & 100   & \multicolumn{1}{c|}{89.4}  & 95.17     & 56.87      & 80.6      & \multicolumn{1}{c|}{99.26}         & 67.54         & 85.51              \\
                                                                                            &                                                                        & ResNet50                            & 99.99 & 100   & \multicolumn{1}{c|}{93.71} & 94.33     & 58.95      & 88.79     & \multicolumn{1}{c|}{97.04}         & 62.35         & 82.51              \\
                                                                                            &                                                                        & ViT                                 & 100   & 100   & \multicolumn{1}{c|}{95.45} & 95.13     & 35.6       & 89.12     & \multicolumn{1}{c|}{98.65}         & 73.79         & 86.61              \\ \cline{2-12} 
                                                                                            & \multirow{8}{*}{Joint}                                
                                                                                                                                                                & MesoNet                             & 92.55 & 97.94 & \multicolumn{1}{c|}{78.42} & 84.01     & 45.87      & 70        & \multicolumn{1}{c|}{56.66}         & 50.57         & 55.03              \\
                                                   &                            & Xception                            & 99.8  & 100   & \multicolumn{1}{c|}{97.62} & 91.56     & 75.49      & 79.38     & \multicolumn{1}{c|}{96.69}         & 63.69         & 78.87              \\                                                           &                                               & MultiAtten                          & 99.38 & 100   & \multicolumn{1}{c|}{78.34} & 92.89     & 58.31      & 77.51     & \multicolumn{1}{c|}{69}            & 58.11         & 60                 \\
                                                                                            &                                                                        & CDCN++                              & 99.99 & 98.89 & \multicolumn{1}{c|}{96.8}  & 78.93     & 61.54      & 76.39     & \multicolumn{1}{c|}{95.79}         & 56.42         & 76.92              \\
                                                                                            &                                                                        & DeepPixel                           & 100   & 100   & \multicolumn{1}{c|}{97.31} & 93.97     & 66.65      & 74.58     & \multicolumn{1}{c|}{99.22}         & 66.36         & 88.45              \\
                                                                                            &                                                                        & ResNet50                            & 99.93 & 100   & \multicolumn{1}{c|}{98.1}  & 93.5      & \textbf{79.64}      & 86.03     & \multicolumn{1}{c|}{98.83}         & 67.18         & 85.55              \\
                                                                                            &                                                                        & ViT                                 & 100   & 100   & \multicolumn{1}{c|}{100}   & 95.85     & 62.74      & 91.88     & \multicolumn{1}{c|}{98.83}         & 75.11         & 88.36              \\
                                                                                            &                                                                        & ViT(shared 8)                       & 100   & 100   & \multicolumn{1}{c|}{99.96} & \textbf{97.99}     & 67.1       & \textbf{95.57}     & \multicolumn{1}{c|}{99.12}         & \textbf{75.9}          & \textbf{89.76}              \\ \hline
\multirow{15}{*}{EER(\%)}                                                                   & \multirow{7}{*}{Sep.}                                                  & MesoNet                             & 11.67 & 0     & \multicolumn{1}{c|}{35.12} & 21.9      & 55.82      & 36.81     & \multicolumn{1}{c|}{38.93}         & 46.16         & 38.78              \\
                                                                                            &                                     & Xception                            & 1.14  & 0     & \multicolumn{1}{c|}{11.31} & 18.57     & 42.3       & 18.13     & \multicolumn{1}{c|}{5.54}          & 38.92         & 24.07              \\                                   
                                                                                            &                                                                        & MultiAtten                          & 2.78  & 0     & \multicolumn{1}{c|}{35.71} & 18.57     & 61.22      & 33.82     & \multicolumn{1}{c|}{35.54}         & 39.73         & 40.46              \\
                                                                                            &                                                                        & CDCN++                              & 0.14  & 3.33  & \multicolumn{1}{c|}{11.9}  & 22.8      & 52.76      & 32.01     & \multicolumn{1}{c|}{7.5}           & 44.75         & 28.19              \\
                                                                                            &                                                                        & DeepPixel                           & 0     & 0     & \multicolumn{1}{c|}{17.86} & 15.71     & 46.42      & 28.23     & \multicolumn{1}{c|}{2.32}          & 36.82         & 22.81              \\
                                                                                            &                                                                        & ResNet50                            & 0.36  & 0     & \multicolumn{1}{c|}{13.69} & 12.38     & 45.24      & 19.94     & \multicolumn{1}{c|}{8.18}          & 40.61         & 26.01              \\
                                                                                            &                                                                        & ViT                                 & 0.14  & 0     & \multicolumn{1}{c|}{11.31} & 11.9      & 60.4       & 19.47     & \multicolumn{1}{c|}{4.64}          & 32.75         & 22.76              \\ \cline{2-12} 
                                                                                            & \multirow{8}{*}{Joint}                                                  & MesoNet                             & 14.8  & 6.67  & \multicolumn{1}{c|}{28.57} & 22.86     & 54.76      & 36.18     & \multicolumn{1}{c|}{44.29}         & 49.81         & 46.94              \\
                                                      &                                       &                                   Xception                            & 1.71  & 0     & \multicolumn{1}{c|}{9.52}  & 17.14     & 22.21      & 28.82     & \multicolumn{1}{c|}{9.29}          & 40.75         & 29.16              \\                                    
                                                                                            &                                                                        & MultiAtten                          & 4.13  & 0     & \multicolumn{1}{c|}{28.57} & 13.81     & 44.42      & 30.32     & \multicolumn{1}{c|}{35.96}         & 43.93         & 43.57              \\
                                                                                            &                                                                        & CDCN++                              & 0.28  & 6.67  & \multicolumn{1}{c|}{10.71} & 31.9      & 43.83      & 33.19     & \multicolumn{1}{c|}{8.39}          & 44.58         & 29.68              \\
                                                                                            &                                                                        & DeepPixel                           & 0.14  & 0     & \multicolumn{1}{c|}{10.12} & 13.33     & 40.19      & 32.8      & \multicolumn{1}{c|}{2.14}          & 37.38         & 19.48              \\
                                                                                            &                                                                        & ResNet50                            & 1.21  & 0     & \multicolumn{1}{c|}{6.55}  & 12.86     & \textbf{28.79}      & 21.55     & \multicolumn{1}{c|}{4.29}          & 36.82         & 23.06              \\
                                                                                            &                                                                        & ViT                                 & 0     & 0     & \multicolumn{1}{c|}{0}     & 9.52      & 46.3       & 16.2      & \multicolumn{1}{c|}{2.68}          & 30.99         & 20.01              \\
                                                                                            &                                                                        & \multicolumn{1}{l|}{ViT(shared 8)} & 0     & 0     & \multicolumn{1}{c|}{0.6}   & \textbf{6.67}      & 43.6       & \textbf{11.64}     & \multicolumn{1}{c|}{2.32}          & \textbf{30.6}          & \textbf{18.8}               \\ \hline
\multirow{15}{*}{\begin{tabular}[c]{@{}c@{}}TPR(\%)@\\ FPR=10\%,\\ TPR(\%)@\\ FPR=1\%\end{tabular}} & \multirow{7}{*}{Sep.}                                                   & MesoNet                             & \multicolumn{3}{c|}{76.73, 34.81}          & \multicolumn{3}{c|}{28.72, 9.1}    & \multicolumn{1}{c|}{23.93, 1.07}   & \multicolumn{2}{c|}{24.92, 2.71}   \\
                                                                                            &                         & Xception                            & \multicolumn{3}{c|}{97.13, 94.2}           & \multicolumn{3}{c|}{55.97, 30.83}  & \multicolumn{1}{c|}{97.68, 84.43}  & \multicolumn{2}{c|}{53.46, 11.19}  \\                                              
                                                                                            &                                                                        & MultiAtten                          & \multicolumn{3}{c|}{93.08, 58.39}          & \multicolumn{3}{c|}{34.13, 13.46}  & \multicolumn{1}{c|}{30.18, 7.68}   & \multicolumn{2}{c|}{32.15, 3.9}    \\
                                                                                            &                                                                        & CDCN++                              & \multicolumn{3}{c|}{98.69, 90.95}          & \multicolumn{3}{c|}{36.51, 18.04}  & \multicolumn{1}{c|}{94.29, 63.75}  & \multicolumn{2}{c|}{51.34, 12.46}  \\
                                                                                            &                                                                        & DeepPixel                           & \multicolumn{3}{c|}{97.07, 93.57}          & \multicolumn{3}{c|}{40.62, 22.09}  & \multicolumn{1}{c|}{98.57, 95.18}  & \multicolumn{2}{c|}{52.42, 9.34}   \\
                                                                                            &                                                                        & ResNet50                            & \multicolumn{3}{c|}{97.63, 94.82}          & \multicolumn{3}{c|}{53.61, 29.16}  & \multicolumn{1}{c|}{94.93, 81.32}  & \multicolumn{2}{c|}{50.43, 8.51}   \\
                                                                                            &                                                                        & ViT                                 & \multicolumn{3}{c|}{98.57, 92.26}          & \multicolumn{3}{c|}{52.69, 29.8}   & \multicolumn{1}{c|}{97.5, 87.86}   & \multicolumn{2}{c|}{56.41, 10.31}  \\ \cline{2-12} 
                                                                                            & \multirow{8}{*}{Joint}                                                           & MesoNet                             & \multicolumn{3}{c|}{72.11, 35.25}          & \multicolumn{3}{c|}{30.66, 10.07}  & \multicolumn{1}{c|}{17.14, 0.54}   & \multicolumn{2}{c|}{19.03, 1.46}   \\                                   &                          
                                                                                       &                                   Xception                            & \multicolumn{3}{c|}{97.38, 93.33}          & \multicolumn{3}{c|}{48.97, 28.75}  & \multicolumn{1}{c|}{91.43, 49.11}  & \multicolumn{2}{c|}{40.18, 5.23}   \\
                                                    &                                      & MultiAtten                          & \multicolumn{3}{c|}{94.88, 71.55}          & \multicolumn{3}{c|}{43.9, 19.84}   & \multicolumn{1}{c|}{25.18, 2.68}   & \multicolumn{2}{c|}{29.86, 2.89}   \\
                                                                                            &                                                                        & CDCN++                              & \multicolumn{3}{c|}{99.31, 95.07}          & \multicolumn{3}{c|}{37.24, 19.59}  & \multicolumn{1}{c|}{93.04, 48.21}  & \multicolumn{2}{c|}{51.62, 15.32}  \\
                                                                                            &                                                                        & DeepPixel                           & \multicolumn{3}{c|}{98.88, 95.57}          & \multicolumn{3}{c|}{40.76, 25.28}  & \multicolumn{1}{c|}{98.04, 96.79}  & \multicolumn{2}{c|}{59.2, 11.93}   \\
                                                                                            &                                                                        & ResNet50                            & \multicolumn{3}{c|}{99.19, 96.51}          & \multicolumn{3}{c|}{58.93, 31.97}  & \multicolumn{1}{c|}{97.68, 86.25}  & \multicolumn{2}{c|}{56.39, 11.48}  \\
                                                                                            &                                                                        & ViT                                 & \multicolumn{3}{c|}{100, 100}              & \multicolumn{3}{c|}{60.18, 32.7}   & \multicolumn{1}{c|}{98.04, 86.62}  & \multicolumn{2}{c|}{63.15, 12.5}   \\
                                                                                            &                                                                        & \multicolumn{1}{l|}{ViT(shared 8)} & \multicolumn{3}{c|}{99.94, 99.38}          & \multicolumn{3}{c|}{\textbf{69.98}, \textbf{36.18}}  & \multicolumn{1}{c|}{98.21, 96.79}  & \multicolumn{2}{c|}{\textbf{66.65}, \textbf{16.85}}  \\ \hline
\end{tabular}}
\label{tab:SOTAappearance}
\vspace{-1.0em}
\end{table*}

\begin{table*}[]
\centering
\caption{Results of rPPG-based unimodal models on the joint face spoofing and forgery detection benchmark. `\_ST' and `\_STW' denote using MSTmap inputs and two-branch (MSTmap+WaveletMap) inputs, respectively.}
\vspace{-0.5em}
\resizebox{0.9\textwidth}{!}{
\begin{tabular}{|c|c|c|cccccc|ccc|}
\hline
\multirow{3}{*}{Metrics}                                                                    & \multirow{3}{*}{\begin{tabular}[c]{@{}c@{}}Sep./\\ Joint\end{tabular}} & \multirow{3}{*}{Method} & \multicolumn{6}{c|}{Face Spoofing Detection}                                    & \multicolumn{3}{c|}{Face Forgery Detection}                             \\ \cline{4-12} 
                                                                                            &                                                                        &                         & \multicolumn{3}{c|}{Intra-testing}         & \multicolumn{3}{c|}{Cross-testing} & \multicolumn{1}{c|}{Intra-testing} & \multicolumn{2}{c|}{Cross-testing} \\ \cline{4-12} 
                                                                                            &                                                                        &                         & SiW   & 3DMAD & \multicolumn{1}{c|}{HKBU}  & MSU       & 3DMask     & ROSE      & \multicolumn{1}{c|}{FF++}          & DFDC          & Celeb-DFv2         \\ \hline
\multirow{11}{*}{AUC(\%)}                                                                   & \multirow{5}{*}{Sep.}                                                  & TransRPPG\_ST           & 88.61 & 92.22 & \multicolumn{1}{c|}{86.24} & 54.19     & 39.9       & 68.98     & \multicolumn{1}{c|}{85.44}         & 53.65         & 60.48              \\
                                                                                            &                                                                        & ResNet18\_ST            & 93.61 & 94.72 & \multicolumn{1}{c|}{94.06} & 57.56     & 31.76      & 70        & \multicolumn{1}{c|}{83.77}         & 54.82         & 63.84              \\
                                                                                            &                                                                        & ResNet18\_STW           & 95.37 & 97.28 & \multicolumn{1}{c|}{94.88} & 58.12     & 38.16      & 71.24     & \multicolumn{1}{c|}{84.56}         & 55.4          & 64.8               \\
                                                                                            &                                                                        & ViT\_ST                 & 96.53 & 96.06 & \multicolumn{1}{c|}{95.77} & 58.66     & 35.35      & 71.97     & \multicolumn{1}{c|}{86.27}         & 54.36         & 63.92              \\
                                                                                            &                                                                        & ViT\_STW                & 97.88 & 98.22 & \multicolumn{1}{c|}{97.02} & 62.31     & 48.13      & \textbf{74.77}     & \multicolumn{1}{c|}{87.41}         & 54.86         & 65.12              \\ \cline{2-12} 
                                                                                            & \multirow{6}{*}{Joint}                                                 & TransRPPG\_ST           & 83.3  & 93.11 & \multicolumn{1}{c|}{80.97} & 56.45     & 68.63      & 66.22     & \multicolumn{1}{c|}{81.67}         & 55.65         & 62.52              \\
                                                                                            &                                                                        & ResNet18\_ST            & 92.66 & 92.22 & \multicolumn{1}{c|}{93.62} & 54.27     & 57.47      & 68.16     & \multicolumn{1}{c|}{82.77}         & 55.83         & 64.19              \\
                                                                                            &                                                                        & ResNet18\_STW           & 94.2  & 95.5  & \multicolumn{1}{c|}{94.9}  & 63.75     & 73.89      & 68.9      & \multicolumn{1}{c|}{86.24}         & \textbf{68.36}         & 66.15              \\
                                                                                            &                                                                        & ViT\_ST                 & 96.56 & 97.83 & \multicolumn{1}{c|}{94.3}  & 56.86     & 62.41      & 67.82     & \multicolumn{1}{c|}{85.53}         & 54.66         & 64.09              \\
                                                                                            &                                                                        & ViT\_STW                & 97.26 & 98.17 & \multicolumn{1}{c|}{96.49} & 64.17     & 74.66      & 72.05     & \multicolumn{1}{c|}{88.81}         & 66.52         & 65.67              \\
                                                                                            &                                                                        & ViT\_STW(shared 8)      & 97.53 & 98.28 & \multicolumn{1}{c|}{96.94} & \textbf{64.72}     & \textbf{79.32}      & 73.94     & \multicolumn{1}{c|}{88.22}         & 64.35         & \textbf{67.2}               \\ \hline
\multirow{11}{*}{EER(\%)}                                                                   & \multirow{5}{*}{Sep.}                                                  & TransRPPG\_ST           & 18.85 & 14.33 & \multicolumn{1}{c|}{22.14} & 50.95     & 59.11      & 36.96     & \multicolumn{1}{c|}{23}            & 46.85         & 39.81              \\
                                                                                            &                                                                        & ResNet18\_ST            & 14.09 & 13.33 & \multicolumn{1}{c|}{13.69} & 46.67     & 65.45      & 35.9      & \multicolumn{1}{c|}{25.71}         & 45.56         & 37.44              \\
                                                                                            &                                                                        & ResNet18\_STW           & 11.53 & 6.67  & \multicolumn{1}{c|}{13.1}  & 43.13     & 60.16      & 35.39     & \multicolumn{1}{c|}{23.46}         & 45.52         & 36.59              \\
                                                                                            &                                                                        & ViT\_ST                 & 9.18  & 13.33 & \multicolumn{1}{c|}{9.52}  & 44.29     & 60.75      & 35        & \multicolumn{1}{c|}{21.25}         & 46.46         & 37.03              \\
                                                                                            &                                                                        & ViT\_STW                & 6.05  & 10    & \multicolumn{1}{c|}{9.52}  & 40        & 51.28      & 32.36     & \multicolumn{1}{c|}{20.43}         & 45.91         & 36.29              \\ \cline{2-12} 
                                                                                            & \multirow{6}{*}{Joint}                                                 & TransRPPG\_ST           & 24.48 & 14.33 & \multicolumn{1}{c|}{26.79} & 48.95     & 37.13      & 39.83     & \multicolumn{1}{c|}{31.61}         & 46.21         & 38.15              \\
                                                                                            &                                                                        & ResNet18\_ST            & 14.31 & 20    & \multicolumn{1}{c|}{13.1}  & 47.14     & 44.65      & 37.71     & \multicolumn{1}{c|}{26.32}         & 45.39         & 36.9               \\
                                                                                            &                                                                        & ResNet18\_STW           & 13.02 & 10    & \multicolumn{1}{c|}{10.71} & 39.05     & 31.73      & 37.59     & \multicolumn{1}{c|}{22.34}         & \textbf{34.13}         & 36.08              \\
                                                                                            &                                                                        & ViT\_ST                 & 9.82  & 10    & \multicolumn{1}{c|}{12.5}  & 47.62     & 41.13      & 38.07     & \multicolumn{1}{c|}{23.21}         & 46.12         & 36.95              \\
                                                                                            &                                                                        & ViT\_STW                & 8.68  & 6.67  & \multicolumn{1}{c|}{8.93}  & \textbf{37.76}     & 30.08      & 36.1      & \multicolumn{1}{c|}{19.78}         & 35.04         & 36.16              \\
                                                                                            &                                                                        & ViT\_STW(shared 8)      & 7.69  & 3.33  & \multicolumn{1}{c|}{10.71} & 40        & \textbf{23.48}      & \textbf{32.25}     & \multicolumn{1}{c|}{22}            & 37.06         & \textbf{35.56}              \\ \hline
\multirow{11}{*}{\begin{tabular}[c]{@{}c@{}}TPR(\%)@\\ FPR=10\%,\\ TPR(\%)@\\ FPR=1\%\end{tabular}} & \multirow{5}{*}{Sep.}                                                  & TransRPPG\_ST           & \multicolumn{3}{c|}{74.27, 31.4}           & \multicolumn{3}{c|}{25.65, 7.8}    & \multicolumn{1}{c|}{68.21, 14.36}  & \multicolumn{2}{c|}{16.4, 1.55}    \\
                                                                                            &                                                                        & ResNet18\_ST            & \multicolumn{3}{c|}{82.35, 36.74}          & \multicolumn{3}{c|}{20.23, 5.36}   & \multicolumn{1}{c|}{63.39, 12.25}  & \multicolumn{2}{c|}{17.68, 1.8}    \\
                                                                                            &                                                                        & ResNet18\_STW           & \multicolumn{3}{c|}{87.59, 53.96}          & \multicolumn{3}{c|}{24.72, 6.41}   & \multicolumn{1}{c|}{69.14, 13.95}  & \multicolumn{2}{c|}{22.57, 2.46}   \\
                                                                                            &                                                                        & ViT\_ST                 & \multicolumn{3}{c|}{91.45, 47.79}          & \multicolumn{3}{c|}{25.22, 8.49}   & \multicolumn{1}{c|}{77.14, 16.96}  & \multicolumn{2}{c|}{16.98, 1.66}   \\
                                                                                            &                                                                        & ViT\_STW                & \multicolumn{3}{c|}{95.26, 58.2}           & \multicolumn{3}{c|}{28.45, 7.52}   & \multicolumn{1}{c|}{80.11, 17.44}  & \multicolumn{2}{c|}{21.62, 2.28}   \\ \cline{2-12} 
                                                                                            & \multirow{6}{*}{Joint}                                                 & TransRPPG\_ST           & \multicolumn{3}{c|}{71.4, 38.09}           & \multicolumn{3}{c|}{34.22, 10.27}  & \multicolumn{1}{c|}{72.32, 18.18}  & \multicolumn{2}{c|}{18.32, 1.74}   \\
                                                                                            &                                                                        & ResNet18\_ST            & \multicolumn{3}{c|}{79.35, 36.62}          & \multicolumn{3}{c|}{27.47, 5.69}   & \multicolumn{1}{c|}{71.96, 17.61}  & \multicolumn{2}{c|}{19.22, 2.1}    \\
                                                                                            &                                                                        & ResNet18\_STW           & \multicolumn{3}{c|}{84.34, 41.98}          & \multicolumn{3}{c|}{32.21, 10.16}  & \multicolumn{1}{c|}{76.64, 17.31}  & \multicolumn{2}{c|}{\textbf{30.4}, \textbf{4.34}}    \\
                                                                                            &                                                                        & ViT\_ST                 & \multicolumn{3}{c|}{89.33, 58.39}          & \multicolumn{3}{c|}{33.63, 9.19}   & \multicolumn{1}{c|}{80.36, 20.18}  & \multicolumn{2}{c|}{17.29, 1.54}   \\
                                                                                            &                                                                        & ViT\_STW                & \multicolumn{3}{c|}{94.45, 50.16}          & \multicolumn{3}{c|}{34.54, 11.02}  & \multicolumn{1}{c|}{81.64, 18.66}  & \multicolumn{2}{c|}{28.7, 3.42}    \\
                                                                                            &                                                                        & ViT\_STW(shared 8)      & \multicolumn{3}{c|}{97.7, 62.28}           & \multicolumn{3}{c|}{\textbf{38.88}, \textbf{12.1}}   & \multicolumn{1}{c|}{84.82, 28.57}  & \multicolumn{2}{c|}{29.29, 3.56}   \\ \hline
\end{tabular}}
\label{tab:SOTArPPG}
\vspace{-0.8em}
\end{table*}

\begin{table*}[]
\centering
\caption{Results of appearance+rPPG based multi-modal models on the joint face spoofing and forgery detection benchmark. `CA', `SE' and `Norm' are short for cross-attention~\cite{yu2022flexible}, squeeze-and-excitation~\cite{casiasurf} and weighted normalization fusion.}
\vspace{-0.5em}
\resizebox{0.94\textwidth}{!}{
\begin{tabular}{|c|c|c|cccccc|ccc|}
\hline
\multirow{3}{*}{Metrics}                                                                    & \multirow{3}{*}{\begin{tabular}[c]{@{}c@{}}Sep./\\ Joint\end{tabular}} & \multirow{3}{*}{Method}     & \multicolumn{6}{c|}{Face Spoofing Detection}                                    & \multicolumn{3}{c|}{Face Forgery Detection}                             \\ \cline{4-12} 
                                                                                            &                                                                        &                             & \multicolumn{3}{c|}{Intra-testing}         & \multicolumn{3}{c|}{Cross-testing} & \multicolumn{1}{c|}{Intra-testing} & \multicolumn{2}{c|}{Cross-testing} \\ \cline{4-12} 
                                                                                            &                                                                        &                             & SiW   & 3DMAD & \multicolumn{1}{c|}{HKBU}  & MSU       & 3DMask     & ROSE      & \multicolumn{1}{c|}{FF++}          & DFDC          & Celeb-DFv2         \\ \hline
\multirow{17}{*}{AUC(\%)}                                                                   & \multirow{8}{*}{Sep.}                                                  & ResNet\_CA                  & 99.97 & 100   & \multicolumn{1}{c|}{92.46} & 95.32     & 65.17      & 87.84     & \multicolumn{1}{c|}{98.52}         & 62.64         & 78.76              \\
                                                                                            &                                                                        & ResNet\_Concat              & 99.97 & 100   & \multicolumn{1}{c|}{95.89} & 93.1      & 58.96      & 88.59     & \multicolumn{1}{c|}{95.86}         & 61.09         & 75.32              \\
                                                                                            &                                                                        & ResNet\_SE\_Concat          & 99.99 & 100   & \multicolumn{1}{c|}{95.64} & 91.88     & 56.72      & 90.65     & \multicolumn{1}{c|}{97.95}         & 64.71         & 76.87              \\
                                                                                            &                                                                        & ResNet\_Norm\_Concat        & 99.99 & 100   & \multicolumn{1}{c|}{97.56} & 94.8      & 55.21      & 91.52     & \multicolumn{1}{c|}{97.36}         & 61.09         & 75.32              \\
                                                                                            &                                                                        & ViT\_CA                     & 99.99 & 100   & \multicolumn{1}{c|}{94.6}  & 90.63     & 39.04      & 85.9      & \multicolumn{1}{c|}{87.28}         & 58.75         & 62.04              \\
                                                                                            &                                                                        & ViT\_Concat                 & 100   & 100   & \multicolumn{1}{c|}{99.72} & 92.03     & 67.35      & 91.94     & \multicolumn{1}{c|}{97.44}         & 62.01         & 67.03              \\
                                                                                            &                                                                        & ViT\_SE\_Concat             & 100   & 100   & \multicolumn{1}{c|}{99.55} & 97.24     & 64.91      & 93.4      & \multicolumn{1}{c|}{97.71}         & 64.76         & 78.52              \\
                                                                                            &                                                                        & ViT\_Norm\_Concat           & 100   & 100   & \multicolumn{1}{c|}{99.95} & 94.33     & 66.76      & \textbf{93.72}     & \multicolumn{1}{c|}{97.44}         & 63.11         & 69.3               \\ \cline{2-12} 
                                                                                            & \multirow{9}{*}{Joint}                                                 & ResNet\_CA                  & 99.96 & 100   & \multicolumn{1}{c|}{99}    & 96.68     & 84.24      & 88.03     & \multicolumn{1}{c|}{98.39}         & \textbf{67.79}         & 78.52              \\
                                                                                            &                                                                        & ResNet\_Concat              & 99.92 & 100   & \multicolumn{1}{c|}{98.03} & 89.88     & 90.58      & 83.07     & \multicolumn{1}{c|}{96.31}         & 62.63         & 76.41              \\
                                                                                            &                                                                        & ResNet\_SE\_Concat          & 99.95 & 100   & \multicolumn{1}{c|}{98.05} & 88.51     & 83.86      & 82.48     & \multicolumn{1}{c|}{96.82}         & 65.6          & 76.2               \\
                                                                                            &                                                                        & ResNet\_Norm\_Concat        & 99.99 & 100   & \multicolumn{1}{c|}{95.73} & 95.33     & 70.76      & 92.07     & \multicolumn{1}{c|}{97.18}         & 68.2          & \textbf{80.9}               \\
                                                                                            &                                                                        & ViT\_CA                     & 95.05 & 98.56 & \multicolumn{1}{c|}{93.85} & 85.8      & 59.11      & 75.34     & \multicolumn{1}{c|}{83.59}         & 55.65         & 60.53              \\
                                                                                            &                                                                        & ViT\_Concat                 & 99.99 & 100   & \multicolumn{1}{c|}{100}   & 96.88     & 78.07      & 89.46     & \multicolumn{1}{c|}{97.85}         & 64.88         & 69.91              \\
                                                                                            &                                                                        & ViT\_SE\_Concat             & 100   & 100   & \multicolumn{1}{c|}{100}   & 94.48     & 81.71      & 90.59     & \multicolumn{1}{c|}{98.21}         & 66.11         & 74.87              \\
                                                                                            &                                                                        & ViT\_Norm\_Concat           & 100   & 100   & \multicolumn{1}{c|}{99.91} & 95.35     & 85.56      & 90.73     & \multicolumn{1}{c|}{97.85}         & 66.25         & 78.57              \\
                                                                                            &                                                                        & ViT\_Norm\_Concat(shared 8) & 100   & 100   & \multicolumn{1}{c|}{99.95} & \textbf{97.88}     & \textbf{88.72}      & 92.5      & \multicolumn{1}{c|}{97.92}         & 67.26         & 79.59              \\ \hline
\multirow{17}{*}{EER(\%)}                                                                   & \multirow{8}{*}{Sep.}                                                  & ResNet\_CA                  & 0.78  & 0     & \multicolumn{1}{c|}{16.07} & 10.95     & 38.9       & 20.49     & \multicolumn{1}{c|}{5.71}          & 41.08         & 27.32              \\
                                                                                            &                                                                        & ResNet\_Concat              & 0.78  & 0     & \multicolumn{1}{c|}{11.31} & 13.81     & 42.77      & 20.06     & \multicolumn{1}{c|}{10}            & 41.63         & 30.71              \\
                                                                                            &                                                                        & ResNet\_SE\_Concat          & 0.36  & 0     & \multicolumn{1}{c|}{11.9}  & 12.86     & 45.12      & 17.89     & \multicolumn{1}{c|}{7.5}           & 38.56         & 29.71              \\
                                                                                            &                                                                        & ResNet\_Norm\_Concat        & 3.6   & 0     & \multicolumn{1}{c|}{8.93}  & 12.48     & 46.18      & 16.71     & \multicolumn{1}{c|}{6.96}          & 41.63         & 30.71              \\
                                                                                            &                                                                        & ViT\_CA                     & 0.64  & 0     & \multicolumn{1}{c|}{13.1}  & 15.24     & 58.28      & 21.79     & \multicolumn{1}{c|}{20.71}         & 44.38         & 38.72              \\
                                                                                            &                                                                        & ViT\_Concat                 & 0     & 0     & \multicolumn{1}{c|}{2.38}  & 17.62     & 38.43      & 16.56     & \multicolumn{1}{c|}{8.21}          & 40.98         & 39.03              \\
                                                                                            &                                                                        & ViT\_SE\_Concat             & 0     & 0     & \multicolumn{1}{c|}{1.19}  & 7.62      & 38.43      & 14.51     & \multicolumn{1}{c|}{7.5}           & 39.99         & 28.64              \\
                                                                                            &                                                                        & ViT\_Norm\_Concat           & 0     & 0     & \multicolumn{1}{c|}{1.19}  & 10.1      & 36.78      & \textbf{13.84}     & \multicolumn{1}{c|}{6.81}          & 39.98         & 36.86              \\ \cline{2-12} 
                                                                                            & \multirow{9}{*}{Joint}                                                 & ResNet\_CA                  & 1     & 0     & \multicolumn{1}{c|}{4.17}  & 10.71     & 24.79      & 20        & \multicolumn{1}{c|}{6.25}          & 36.48         & 28.61              \\
                                                                                            &                                                                        & ResNet\_Concat              & 1.57  & 0     & \multicolumn{1}{c|}{7.14}  & 17.14     & 17.86      & 23.84     & \multicolumn{1}{c|}{8.57}          & 40.09         & 30.4               \\
                                                                                            &                                                                        & ResNet\_SE\_Concat          & 1.28  & 0     & \multicolumn{1}{c|}{5.36}  & 18.1      & 24.09      & 25.32     & \multicolumn{1}{c|}{9.82}          & 36.92         & 28.05              \\
                                                                                            &                                                                        & ResNet\_Norm\_Concat        & 0.57  & 0     & \multicolumn{1}{c|}{11.9}  & 9.1       & 32.78      & 14.84     & \multicolumn{1}{c|}{7.09}          & 36.29         & \textbf{26.25}              \\
                                                                                            &                                                                        & ViT\_CA                     & 12.74 & 3.33  & \multicolumn{1}{c|}{14.88} & 19.52     & 43.36      & 32.01     & \multicolumn{1}{c|}{23.39}         & 44.55         & 40.63              \\
                                                                                            &                                                                        & ViT\_Concat                 & 0.28  & 0     & \multicolumn{1}{c|}{0.6}   & 10        & 27.61      & 18.17     & \multicolumn{1}{c|}{6.61}          & 39.31         & 36.46              \\
                                                                                            &                                                                        & ViT\_SE\_Concat             & 0     & 0     & \multicolumn{1}{c|}{0.6}   & 13.81     & 25.38      & 16.71     & \multicolumn{1}{c|}{6.25}          & 38.96         & 30.2               \\
                                                                                            &                                                                        & ViT\_Norm\_Concat           & 0     & 0     & \multicolumn{1}{c|}{0.6}   & 12.38     & 21.62      & 17.38     & \multicolumn{1}{c|}{6.61}          & 37.81         & 30.24              \\
                                                                                            &                                                                        & ViT\_Norm\_Concat(shared 8) & 0     & 0     & \multicolumn{1}{c|}{0}     & \textbf{5.26}      & \textbf{18.83}      & 15.49     & \multicolumn{1}{c|}{6.5}           & \textbf{36.22}         & 27.1               \\ \hline
\multirow{17}{*}{\begin{tabular}[c]{@{}c@{}}TPR(\%)@\\ FPR=10\%,\\ TPR(\%)@\\ FPR=1\%\end{tabular}} & \multirow{8}{*}{Sep.}                                                  & ResNet\_CA                  & \multicolumn{3}{c|}{97.75, 93.7}           & \multicolumn{3}{c|}{55.88, 26.72}  & \multicolumn{1}{c|}{96.79, 79.82}  & \multicolumn{2}{c|}{42.06, 5.44}   \\
                                                                                            &                                                                        & ResNet\_Concat              & \multicolumn{3}{c|}{98.63, 94.76}          & \multicolumn{3}{c|}{54.33, 29.62}  & \multicolumn{1}{c|}{90, 50.71}     & \multicolumn{2}{c|}{39.44, 4.85}   \\
                                                                                            &                                                                        & ResNet\_SE\_Concat          & \multicolumn{3}{c|}{98.81, 95.82}          & \multicolumn{3}{c|}{55.47, 24.5}   & \multicolumn{1}{c|}{94.46, 81.07}  & \multicolumn{2}{c|}{35.91, 4.39}   \\
                                                                                            &                                                                        & ResNet\_Norm\_Concat        & \multicolumn{3}{c|}{98.62, 92.57}          & \multicolumn{3}{c|}{62.53, 30.25}  & \multicolumn{1}{c|}{94, 74.71}     & \multicolumn{2}{c|}{44.48, 8.85}   \\
                                                                                            &                                                                        & ViT\_CA                     & \multicolumn{3}{c|}{99.19, 94.26}          & \multicolumn{3}{c|}{46.25, 20.98}  & \multicolumn{1}{c|}{73.04, 15.75}  & \multicolumn{2}{c|}{28.67, 2.7}    \\
                                                                                            &                                                                        & ViT\_Concat                 & \multicolumn{3}{c|}{100, 99}               & \multicolumn{3}{c|}{64.26, 32.05}  & \multicolumn{1}{c|}{92.32, 61.43}  & \multicolumn{2}{c|}{29.37, 3.6}    \\
                                                                                            &                                                                        & ViT\_SE\_Concat             & \multicolumn{3}{c|}{100, 99}               & \multicolumn{3}{c|}{65.17, 32.84}  & \multicolumn{1}{c|}{93.39, 76.96}  & \multicolumn{2}{c|}{34.91, 4.13}   \\
                                                                                            &                                                                        & ViT\_Norm\_Concat           & \multicolumn{3}{c|}{100, 99.56}            & \multicolumn{3}{c|}{66.32, 33.5}   & \multicolumn{1}{c|}{94.32, 71.43}  & \multicolumn{2}{c|}{36.81, 6.3}    \\ \cline{2-12} 
                                                                                            & \multirow{9}{*}{Joint}                                                 & ResNet\_CA                  & \multicolumn{3}{c|}{99.81, 97.38}          & \multicolumn{3}{c|}{66.10, 32.04}  & \multicolumn{1}{c|}{96.61, 83.04}  & \multicolumn{2}{c|}{47.04, 7.18}   \\
                                                                                            &                                                                        & ResNet\_Concat              & \multicolumn{3}{c|}{99.25, 94.95}          & \multicolumn{3}{c|}{57.58, 30.01}  & \multicolumn{1}{c|}{92.5, 74.64}   & \multicolumn{2}{c|}{42.33, 6.35}   \\
                                                                                            &                                                                        & ResNet\_SE\_Concat          & \multicolumn{3}{c|}{99.75, 95.07}          & \multicolumn{3}{c|}{52.58, 18.23}  & \multicolumn{1}{c|}{90.18, 79.29}  & \multicolumn{2}{c|}{42.23, 8.75}   \\
                                                                                            &                                                                        & ResNet\_Norm\_Concat        & \multicolumn{3}{c|}{99.19, 92.58}          & \multicolumn{3}{c|}{66.96, 32.67}  & \multicolumn{1}{c|}{93.71, 72.12}  & \multicolumn{2}{c|}{51.34, \textbf{11.15}}  \\
                                                                                            &                                                                        & ViT\_CA                     & \multicolumn{3}{c|}{83.47, 48.78}          & \multicolumn{3}{c|}{35.94, 13.3}   & \multicolumn{1}{c|}{60.89, 10.21}  & \multicolumn{2}{c|}{27.79, 2.64}   \\
                                                                                            &                                                                        & ViT\_Concat                 & \multicolumn{3}{c|}{100, 99.75}            & \multicolumn{3}{c|}{65.84, 33.27}  & \multicolumn{1}{c|}{93.75, 81.61}  & \multicolumn{2}{c|}{36.1, 6.01}    \\
                                                                                            &                                                                        & ViT\_SE\_Concat             & \multicolumn{3}{c|}{100, 100}              & \multicolumn{3}{c|}{69.42, 32.57}  & \multicolumn{1}{c|}{94.46, 79.11}  & \multicolumn{2}{c|}{38.46, 5.73}   \\
                                                                                            &                                                                        & ViT\_Norm\_Concat           & \multicolumn{3}{c|}{100, 99.81}            & \multicolumn{3}{c|}{71.23, 35.96}  & \multicolumn{1}{c|}{94.93, 84.82}  & \multicolumn{2}{c|}{40.82, 8.32}   \\
                                                                                            &                                                                        & ViT\_Norm\_Concat(shared 8) & \multicolumn{3}{c|}{100, 99.89}            & \multicolumn{3}{c|}{\textbf{73.83}, \textbf{41.04}}  & \multicolumn{1}{c|}{94.28, 84.24}  & \multicolumn{2}{c|}{\textbf{44.39}, 9.59}   \\ \hline
\end{tabular}
}
\label{tab:fusion}
\end{table*}

 \vspace{0.1em}
 \textbf{Protocols.}\quad  To fairly evaluate the efficacy of the benchmark baselines, we design the protocols under two perspectives: 1) task concurrence; and 2) generalization to domain shifts. As for the former one, separate training on each single task vs. joint training within both two tasks is conducted for verifying the concurrence of face spoofing and forgery detection tasks as well as the effectiveness of multi-task learning. In terms of the latter one, both intra- and cross-domain testings are performed to evaluate the generalization capacities of models. 
 
 • Separate training. This protocol is implemented via training two separate models on datasets $\{\text{SiW, 3DMAD, HKBU}\}$ and $\{\text{FF++}\}$ for face spoofing and forgery detection tasks, respectively. The trained models are tested on both intra-domain (i.e., $\{\text{SiW, 3DMAD, HKBU}\}$ and $\{\text{FF++}\}$) and cross-domain (i.e., $\{\text{MSU, 3DMask, ROSE}\}$ and $\{\text{DFDC, CelebDFv2}\}$) scenarios for two tasks, respectively.
 
• Joint training. Instead of training separate models, the joint training protocol is to train a unified model for two tasks via leveraging all data in $\{\text{SiW, 3DMAD, HKBU, FF++}\}$. Similarly, the trained unified model is evaluated on both intra-domain and cross-domain scenarios for two tasks.
 
Note that the cross-domain testings for both separate and joint protocols are more challenging as the testing set covers more unseen datasets and more complex unknown attacks, which are correlated to real-world scenarios.   

\vspace{0.1em}
 \textbf{Evaluation metrics.}\quad  For all experiments, Area Under Curve (AUC) and Equal Error Rate (EER)~\cite{ACER} are utilized for performance evaluation. Besides, the evaluations under intra- and cross-domain scenarios among multiple (merged) datasets are investigated by using the metric of single-side True Positive Rate (TPR)@False Positive Rate (FPR)~\cite{wang2022domain}, which is more suitable for realistic spectacles.

\vspace{-1.2em}
\section{Experiments}
\vspace{-0.2em}
\label{sec:experiment}



\begin{figure*}[t]
\centering
\includegraphics[scale=0.6]{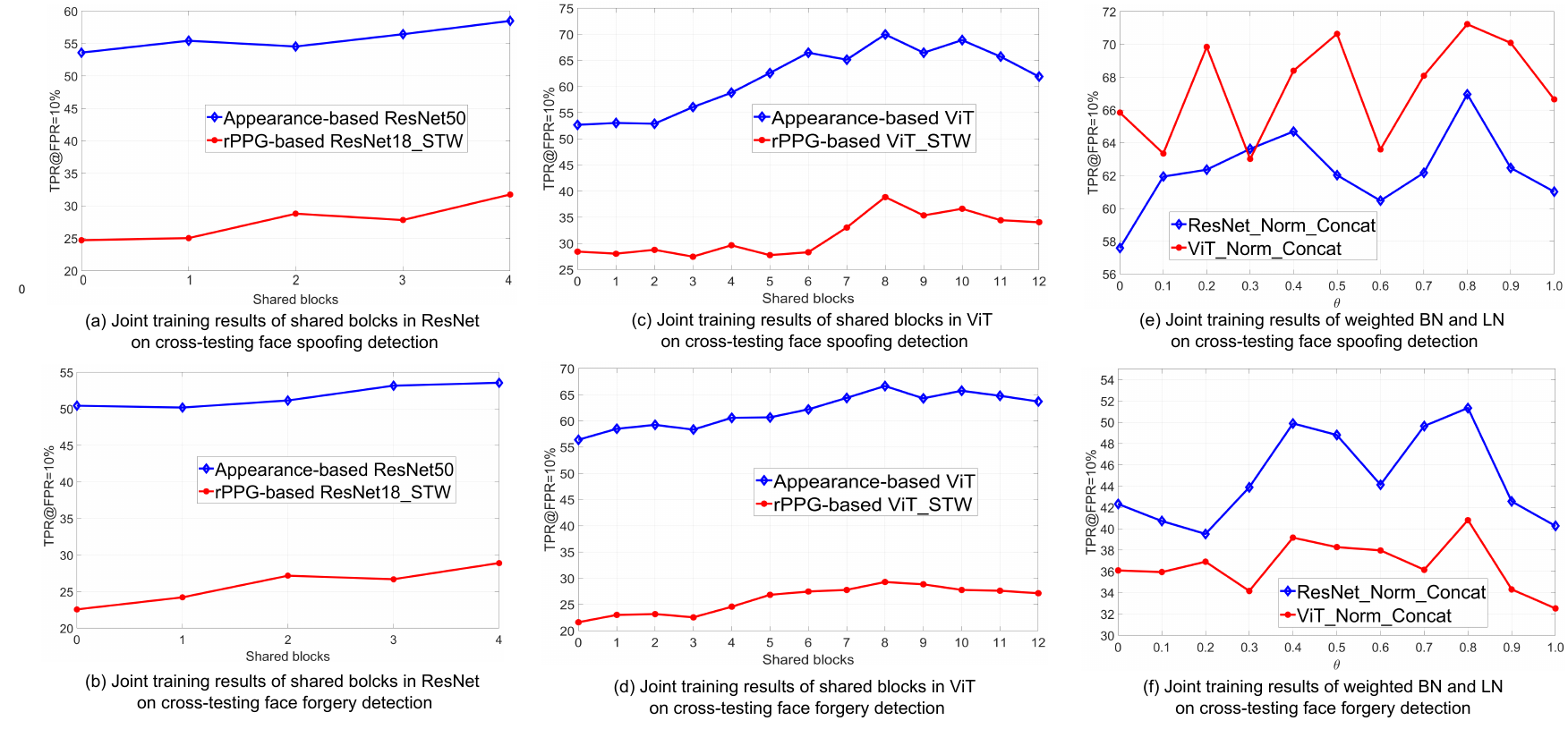}
\vspace{-0.6em}
 \caption{\small{
  Ablation studies of (a)-(d) the joint training architectures with different shared blocks using 2 heads 2 classes setting; and (e)(f) $\theta$ in the weighted normalization fusion.}
  }
 
\label{fig:Ablation12}
\end{figure*}

\begin{table*}[]
\centering
\caption{Ablation study results of the joint training architectures.}
\vspace{-0.5em}
\resizebox{0.94\textwidth}{!}{
\begin{tabular}{|c|c|c|cccccc|ccc|}
\hline
\multirow{3}{*}{Modality}                                                     & \multirow{3}{*}{\begin{tabular}[c]{@{}c@{}}Multi-task\\ Prediection\\ Head\end{tabular}} & \multirow{3}{*}{Method} & \multicolumn{6}{c|}{Face Spoofing Detection  (AUC(\%))}                                       & \multicolumn{3}{c|}{Face Forgery Detection  (AUC(\%))}                  \\ \cline{4-12} 
                                                                              &                                                                                          &                         & \multicolumn{3}{c|}{Intra-testing}         & \multicolumn{3}{c|}{Cross-testing}               & \multicolumn{1}{c|}{Intra-testing} & \multicolumn{2}{c|}{Cross-testing} \\ \cline{4-12} 
                                                                              &                                                                                          &                         & SiW   & 3DMAD & \multicolumn{1}{c|}{HKBU}  & MSU            & 3DMask         & ROSE           & \multicolumn{1}{c|}{FF++}          & DFDC             & Celeb-DFv2      \\ \hline
\multirow{8}{*}{\begin{tabular}[c]{@{}c@{}}Visual\\ Appearance\end{tabular}}  & \multirow{2}{*}{Separate training}                                                                & ResNet50                & 99.99 & 100   & \multicolumn{1}{c|}{93.71} & 94.33          & 58.95          & 88.79          & \multicolumn{1}{c|}{97.04}         & 62.35            & 82.51           \\
                                                                              &                                                                                          & ViT                     & 100   & 100   & \multicolumn{1}{c|}{95.45} & 95.13          & 35.6           & 89.12          & \multicolumn{1}{c|}{98.65}         & 73.79            & 86.61           \\ \cline{2-12} 
                                                                              & \multirow{2}{*}{\begin{tabular}[c]{@{}c@{}}Joint with \\ 1 head 2 classes\end{tabular}}  & ResNet50                & 99.93 & 100   & \multicolumn{1}{c|}{98.1}  & 93.5           & \textbf{79.64} & 86.03          & \multicolumn{1}{c|}{98.83}         & 67.18            & 85.55           \\
                                                                              &                                                                                          & ViT                     & 100   & 100   & \multicolumn{1}{c|}{100}   & 95.85          & 62.74          & 91.88          & \multicolumn{1}{c|}{98.83}         & 75.11            & 88.36           \\ \cline{2-12} 
                                                                              & \multirow{2}{*}{\begin{tabular}[c]{@{}c@{}}Joint with \\ 2 heads 2 classes\end{tabular}} & ResNet50                & 99.99 & 100   & \multicolumn{1}{c|}{98.13} & \textbf{95.95} & 56.51          & 90.68          & \multicolumn{1}{c|}{98.96}         & 64.26            & 86.2            \\
                                                                              &                                                                                          & ViT                     & 100   & 100   & \multicolumn{1}{c|}{99.4}  & 95.29          & 60.6           & \textbf{92.3}  & \multicolumn{1}{c|}{98.87}         & \textbf{75.72}   & \textbf{89.28}  \\ \cline{2-12} 
                                                                              & \multirow{2}{*}{\begin{tabular}[c]{@{}c@{}}Joint with \\ 1 head 3 classes\end{tabular}}  & ResNet50                & 99.99 & 100   & \multicolumn{1}{c|}{97.19} & 93.29          & 54.67          & 82.13          & \multicolumn{1}{c|}{98.51}         & 66.1             & 88.04           \\
                                                                              &                                                                                          & ViT                     & 100   & 100   & \multicolumn{1}{c|}{99.51} & 92.97          & 62.7           & 86.68          & \multicolumn{1}{c|}{98.15}         & 74.82            & 88.65           \\ \hline \hline
\multirow{8}{*}{\begin{tabular}[c]{@{}c@{}}Physiological\\ rPPG\end{tabular}} & \multirow{2}{*}{Separate training}                                                                & ResNet18\_STW           & 95.37 & 97.28 & \multicolumn{1}{c|}{94.88} & 58.12          & 38.16          & 71.24          & \multicolumn{1}{c|}{84.56}         & 55.4             & 64.8            \\
                                                                              &                                                                                          & ViT\_STW                & 97.88 & 98.22 & \multicolumn{1}{c|}{97.02} & 62.31          & 48.13          & \textbf{74.77} & \multicolumn{1}{c|}{87.41}         & 54.86            & 65.12           \\ \cline{2-12} 
                                                                              & \multirow{2}{*}{\begin{tabular}[c]{@{}c@{}}Joint with \\ 1 head 2 classes\end{tabular}}  & ResNet18\_STW           & 94.2  & 95.5  & \multicolumn{1}{c|}{94.9}  & 63.75          & 73.89          & 68.9           & \multicolumn{1}{c|}{86.24}         & \textbf{68.36}   & \textbf{66.15}  \\
                                                                              &                                                                                          & ViT\_STW                & 97.26 & 98.17 & \multicolumn{1}{c|}{96.49} & \textbf{64.17} & \textbf{74.66} & 72.05          & \multicolumn{1}{c|}{88.81}         & 66.52            & 65.67           \\ \cline{2-12} 
                                                                              & \multirow{2}{*}{\begin{tabular}[c]{@{}c@{}}Joint with \\ 2 heads 2 classes\end{tabular}} & ResNet18\_STW           & 94.76 & 95.78 & \multicolumn{1}{c|}{96.58} & 58.6           & 69.55          & 69.45          & \multicolumn{1}{c|}{84.89}         & 61.35            & 62.3            \\
                                                                              &                                                                                          & ViT\_STW                & 97.67 & 98.67 & \multicolumn{1}{c|}{98.01} & 63.86          & 67.84          & 72.22          & \multicolumn{1}{c|}{85.9}          & 59.83            & 64.28           \\ \cline{2-12} 
                                                                              & \multirow{2}{*}{\begin{tabular}[c]{@{}c@{}}Joint with \\ 1 head 3 classes\end{tabular}}  & ResNet18\_STW           & 95.38 & 97.44 & \multicolumn{1}{c|}{94.29} & 60.52          & 64.85          & 66.2           & \multicolumn{1}{c|}{84.82}         & 66.61            & 63.46           \\
                                                                              &                                                                                          & ViT\_STW                & 97.9  & 99.17 & \multicolumn{1}{c|}{97.25} & 63.57          & 64.01          & 72.38          & \multicolumn{1}{c|}{84.04}         & 61.49            & 62.19           \\ \hline
\end{tabular}}
\label{tab:heads}
\end{table*}

\begin{table*}[]
\centering
\caption{Ablation study results of joint training with different sampling strategies.}
\vspace{-0.5em}
\resizebox{0.9\textwidth}{!}{
\begin{tabular}{|c|c|c|cccccc|ccc|}
\hline
\multirow{3}{*}{Modality}                                                     & \multirow{3}{*}{\begin{tabular}[c]{@{}c@{}}Sampling\\ Strategy\end{tabular}} & \multirow{3}{*}{Method} & \multicolumn{6}{c|}{Face Spoofing Detection  (AUC(\%))}                                       & \multicolumn{3}{c|}{Face Forgery Detection  (AUC(\%))}                  \\ \cline{4-12} 
                                                                              &                                                                              &                         & \multicolumn{3}{c|}{Intra-testing}         & \multicolumn{3}{c|}{Cross-testing}               & \multicolumn{1}{c|}{Intra-testing} & \multicolumn{2}{c|}{Cross-testing} \\ \cline{4-12} 
                                                                              &                                                                              &                         & SiW   & 3DMAD & \multicolumn{1}{c|}{HKBU}  & MSU            & 3DMask         & ROSE           & \multicolumn{1}{c|}{FF++}          & DFDC             & Celeb-DFv2      \\ \hline
\multirow{8}{*}{\begin{tabular}[c]{@{}c@{}}Visual\\ Appearance\end{tabular}}  & \multirow{2}{*}{Random}                                                      & ResNet50                & 99.93 & 100   & \multicolumn{1}{c|}{98.1}  & 93.5           & \textbf{79.64} & 86.03          & \multicolumn{1}{c|}{98.83}         & 67.18            & 85.55           \\
                                                                              &                                                                              & ViT                     & 100   & 100   & \multicolumn{1}{c|}{100}   & 95.85          & 62.74          & \textbf{91.88} & \multicolumn{1}{c|}{98.83}         & \textbf{75.11}   & \textbf{88.36}  \\ \cline{2-12} 
                                                                              & \multirow{2}{*}{Simulteanous}                                                & ResNet50                & 99.91 & 100   & \multicolumn{1}{c|}{97.93} & 95.78          & 55.81          & 84.78          & \multicolumn{1}{c|}{98.19}         & 65.61            & 84.93           \\
                                                                              &                                                                              & ViT                     & 100   & 100   & \multicolumn{1}{c|}{100}   & 95.14          & 73.89          & 88.46          & \multicolumn{1}{c|}{98.63}         & 74.5             & 87.29           \\ \cline{2-12} 
                                                                              & \multirow{2}{*}{Alternating}                                                 & ResNet50                & 99.91 & 100   & \multicolumn{1}{c|}{97.7}  & 88.84          & 70.1           & 75.39          & \multicolumn{1}{c|}{97.26}         & 66.1             & 78.59           \\
                                                                              &                                                                              & ViT                     & 100   & 100   & \multicolumn{1}{c|}{100}   & \textbf{96.29} & 51.11          & 88.39          & \multicolumn{1}{c|}{98.87}         & 73.55            & 87.89           \\ \cline{2-12} 
                                                                              & \multirow{2}{*}{Task-by-Task}                                                & ResNet50                & 99.7  & 100   & \multicolumn{1}{c|}{88.4}  & 81.44          & 71.37          & 77.25          & \multicolumn{1}{c|}{98.18}         & 63.21            & 79.94           \\
                                                                              &                                                                              & ViT                     & 100   & 100   & \multicolumn{1}{c|}{99.91} & 95.68          & 65.98          & 84.27          & \multicolumn{1}{c|}{98}            & 73.45            & 82.74           \\  \hline \hline
\multirow{8}{*}{\begin{tabular}[c]{@{}c@{}}Physiological\\ rPPG\end{tabular}} & \multirow{2}{*}{Random}                                                      & ResNet18\_STW           & 94.2  & 95.5  & \multicolumn{1}{c|}{94.9}  & 63.75          & 73.89          & 68.9           & \multicolumn{1}{c|}{86.24}         & \textbf{68.36}   & 66.15           \\
                                                                              &                                                                              & ViT\_STW                & 97.26 & 98.17 & \multicolumn{1}{c|}{96.49} & \textbf{64.17} & 74.66          & \textbf{72.05} & \multicolumn{1}{c|}{88.81}         & 66.52            & 65.67           \\ \cline{2-12} 
                                                                              & \multirow{2}{*}{Simulteanous}                                                & ResNet18\_STW           & 94.9  & 96.33 & \multicolumn{1}{c|}{95.83} & 60.1           & 68.19          & 68.57          & \multicolumn{1}{c|}{80.24}         & 60.64            & 67.35           \\
                                                                              &                                                                              & ViT\_STW                & 97.57 & 98.56 & \multicolumn{1}{c|}{97.06} & 61.02          & 72.14          & 69.54          & \multicolumn{1}{c|}{85.6}          & 59.82            & \textbf{68.43}  \\ \cline{2-12} 
                                                                              & \multirow{2}{*}{Alternating}                                                 & ResNet18\_STW           & 92.65 & 92.94 & \multicolumn{1}{c|}{93.63} & 61.78          & 73.01          & 64.81          & \multicolumn{1}{c|}{76.22}         & 59.81            & 63.48           \\
                                                                              &                                                                              & ViT\_STW                & 97.13 & 96.94 & \multicolumn{1}{c|}{96.13} & 60.82          & 61.65          & 71.39          & \multicolumn{1}{c|}{83.41}         & 59.87            & 66.15           \\ \cline{2-12} 
                                                                              & \multirow{2}{*}{Task-by-Task}                                                & ResNet18\_STW           & 84.89 & 94.72 & \multicolumn{1}{c|}{86.63} & 53.8           & \textbf{86.22} & 59.58          & \multicolumn{1}{c|}{80.82}         & 55.52            & 59.53           \\
                                                                              &                                                                              & ViT\_STW                & 96.9  & 99.28 & \multicolumn{1}{c|}{96.06} & 65             & 51.5           & 71.86          & \multicolumn{1}{c|}{85.36}         & 54.05            & 56.8            \\ \hline
\end{tabular}}
\label{tab:sampling}
\vspace{-1.0em}
\end{table*}

\begin{table*}[]
\centering
\caption{Ablation study results of joint training with extra bonafides and attacks.}
\vspace{-0.5em}
\resizebox{0.9\textwidth}{!}{
\begin{tabular}{|c|c|c|cccccc|ccc|}
\hline
\multirow{3}{*}{Modality}                                                     & \multirow{3}{*}{\begin{tabular}[c]{@{}c@{}}Extra\\ Bonafide/\\ Attack\end{tabular}}                        & \multirow{3}{*}{Method} & \multicolumn{6}{c|}{Face Spoofing Detection  (AUC(\%))}                                       & \multicolumn{3}{c|}{Face Forgery Detection  (AUC(\%))}                   \\ \cline{4-12} 
                                                                              &                                                                                                            &                         & \multicolumn{3}{c|}{Intra-testing}         & \multicolumn{3}{c|}{Cross-testing}               & \multicolumn{1}{c|}{Intra-testing}  & \multicolumn{2}{c|}{Cross-testing} \\ \cline{4-12} 
                                                                              &                                                                                                            &                         & SiW   & 3DMAD & \multicolumn{1}{c|}{HKBU}  & MSU            & 3DMask         & ROSE           & \multicolumn{1}{c|}{FF++}           & DFDC             & Celeb-DFv2      \\ \hline
\multirow{8}{*}{\begin{tabular}[c]{@{}c@{}}Visual\\ Appearance\end{tabular}}  & \multirow{2}{*}{\begin{tabular}[c]{@{}c@{}}-\\ (Separate)\end{tabular}}                                    & ResNet50                & 99.99 & 100   & \multicolumn{1}{c|}{93.71} & 94.33          & 58.95          & 88.79          & \multicolumn{1}{c|}{97.04}          & 62.35            & 82.51           \\
                                                                              &                                                                                                            & ViT                     & 100   & 100   & \multicolumn{1}{c|}{95.45} & 95.13          & 35.6           & 89.12          & \multicolumn{1}{c|}{98.65}          & 73.79            & 86.61           \\ \cline{2-12} 
                                                                              & \multicolumn{1}{l|}{\multirow{2}{*}{\begin{tabular}[c]{@{}l@{}}Both bonafides\\ and attacks\end{tabular}}} & ResNet50                & 99.93 & 100   & \multicolumn{1}{c|}{98.1}  & 93.5           & \textbf{79.64} & 86.03          & \multicolumn{1}{c|}{98.83}          & 67.18            & 85.55           \\
                                                                              & \multicolumn{1}{l|}{}                                                                                      & ViT                     & 100   & 100   & \multicolumn{1}{c|}{100}   & 95.85          & 62.74          & \textbf{91.88} & \multicolumn{1}{c|}{98.83}          & \textbf{75.11}   & \textbf{88.36}  \\ \cline{2-12} 
                                                                              & \multirow{2}{*}{Only bonafides}                                                                            & ResNet50                & 100   & 100   & \multicolumn{1}{c|}{92.52} & 93.17          & 59.23          & 87.21          & \multicolumn{1}{c|}{98.62}          & 67.21            & 86.19           \\
                                                                              &                                                                                                            & ViT                     & 100   & 100   & \multicolumn{1}{c|}{98.02} & \textbf{96.23} & 47.34          & 86.51          & \multicolumn{1}{c|}{98.87} & 75.4             & 90.2            \\ \cline{2-12} 
                                                                              & \multirow{2}{*}{Only attacks}                                                                              & ResNet50                & 99.98 & 100   & \multicolumn{1}{c|}{99.72} & 95.56          & 68.26          & 81.75          & \multicolumn{1}{c|}{98.81}          & 61.12            & 85.65           \\
                                                                              &                                                                                                            & ViT                     & 100   & 100   & \multicolumn{1}{c|}{100}   & 93.23          & 54.26          & 90.88          & \multicolumn{1}{c|}{90.1}           & 68.07            & 78.28           \\ \hline \hline
\multirow{8}{*}{\begin{tabular}[c]{@{}c@{}}Physiological\\ rPPG\end{tabular}} & \multirow{2}{*}{\begin{tabular}[c]{@{}c@{}}-\\ (Separate)\end{tabular}}                                    & ResNet18\_STW           & 95.37 & 97.28 & \multicolumn{1}{c|}{94.88} & 58.12          & 38.16          & 71.24          & \multicolumn{1}{c|}{84.56}          & 55.4             & 64.8            \\
                                                                              &                                                                                                            & ViT\_STW                & 97.88 & 98.22 & \multicolumn{1}{c|}{97.02} & 62.31          & 48.13          & 74.77          & \multicolumn{1}{c|}{87.41}          & 54.86            & 65.12           \\ \cline{2-12} 
                                                                              & \multirow{2}{*}{\begin{tabular}[c]{@{}c@{}}Both bonafides\\ and attacks\end{tabular}}                      & ResNet18\_STW           & 94.2  & 95.5  & \multicolumn{1}{c|}{94.9}  & \textbf{63.75} & \textbf{73.89} & \textbf{68.9}  & \multicolumn{1}{c|}{86.24}          & \textbf{68.36}   & 66.15           \\
                                                                              &                                                                                                            & ViT\_STW                & 97.26 & 98.17 & \multicolumn{1}{c|}{96.49} & \textbf{64.17} & \textbf{74.66} & \textbf{72.05} & \multicolumn{1}{c|}{88.81}          & 66.52            & 65.67           \\ \cline{2-12} 
                                                                              & \multirow{2}{*}{Only bonafides}                                                                            & ResNet18\_STW           & 95.86 & 98    & \multicolumn{1}{c|}{93.99} & 55.34          & 64.87          & 57.07          & \multicolumn{1}{c|}{82.72}          & 67.63            & \textbf{66.18}  \\
                                                                              &                                                                                                            & ViT\_STW                & 89.35 & 96.06 & \multicolumn{1}{c|}{91.71} & 55.28          & 30.94          & 69.06          & \multicolumn{1}{c|}{85.77}          & 66.06            & 63.15           \\ \cline{2-12} 
                                                                              & \multirow{2}{*}{Only attacks}                                                                              & ResNet18\_STW           & 95.59 & 97.33 & \multicolumn{1}{c|}{96.27} & 55.33          & 52.81          & 71.82          & \multicolumn{1}{c|}{83.06}          & 58.4             & 62.78           \\
                                                                              &                                                                                                            & ViT\_STW                & 89.21 & 94.83 & \multicolumn{1}{c|}{93.06} & 55.1           & 51.52          & 70.7           & \multicolumn{1}{c|}{86.46}          & 56.94            & 62.25           \\ \hline
\end{tabular}}
\label{tab:extraPartial}
\end{table*}

\begin{table*}[]
\centering
\caption{Ablation study results of joint training with different face analysis tasks on RAF-DB~\cite{li2019reliable} dataset. The two results in the column `Extra Task Accuracy' denote joint training accuracy on FAS and face forgery detection, respectively.}
\vspace{-0.5em}
\resizebox{0.9\textwidth}{!}{
\begin{tabular}{|c|c|c|cccccc|ccc|}
\hline
\multirow{3}{*}{Joint Task}                                                           & \multirow{3}{*}{Method} & \multirow{3}{*}{\begin{tabular}[c]{@{}c@{}}Extra\\ Task\\  Accuracy(\%)\end{tabular}} & \multicolumn{6}{c|}{Face Spoofing Detection  (AUC(\%))}                                         & \multicolumn{3}{c|}{Face Forgery Detection  (AUC(\%))}                    \\ \cline{4-12} 
                                                                                      &                         &                                                                                       & \multicolumn{3}{c|}{Intra-testing}         & \multicolumn{3}{c|}{Cross-testing}               & \multicolumn{1}{c|}{Intra-testing} & \multicolumn{2}{c|}{Cross-testing} \\ \cline{4-12} 
                                                                                      &                         &                                                                                       & SiW   & 3DMAD & \multicolumn{1}{c|}{HKBU}  & MSU            & 3DMask         & ROSE           & \multicolumn{1}{c|}{FF++}          & DFDC             & Celeb-DFv2      \\ \hline
\multirow{2}{*}{\begin{tabular}[c]{@{}c@{}}Separate\\ (non-joint)\end{tabular}}       & ResNet50                & -                                                                                     & 99.99 & 100   & \multicolumn{1}{c|}{93.71} & 94.33          & 58.95          & 88.79          & \multicolumn{1}{c|}{97.04}         & 62.35            & 82.51           \\
                                                                                      & ViT                     & -                                                                                     & 100   & 100   & \multicolumn{1}{c|}{95.45} & 95.13          & 35.6           & 89.12          & \multicolumn{1}{c|}{98.65}         & 73.79            & 86.61           \\ \hline
\multirow{2}{*}{\begin{tabular}[c]{@{}c@{}}FAS and Forgery \\ Detection\end{tabular}} & ResNet50                & -                                                                                     & 99.93 & 100   & \multicolumn{1}{c|}{98.1}  & 93.5           & \textbf{79.64} & 86.03          & \multicolumn{1}{c|}{98.83}         & 67.18            & 85.55           \\
                                                                                      & ViT                     & -                                                                                     & 100   & 100   & \multicolumn{1}{c|}{100}   & 95.85          & 62.74          & \textbf{91.88} & \multicolumn{1}{c|}{98.83}         & \textbf{75.11}   & \textbf{88.36}  \\ \hline
\multirow{2}{*}{\begin{tabular}[c]{@{}c@{}}Expression \\ Recognition\end{tabular}}    & ResNet50                & \multicolumn{1}{l|}{83.05, 81.71}                                                     & 99.98 & 100   & \multicolumn{1}{c|}{91.26} & \textbf{96.02} & 44.19          & 82.83          & \multicolumn{1}{c|}{97.87}         & 67.72            & 84.6            \\
                                                                                      & ViT                     & \multicolumn{1}{l|}{87.32, 87.52}                                                     & 100   & 100   & \multicolumn{1}{c|}{95.85} & 95.5           & 36.18          & 88.4           & \multicolumn{1}{c|}{98.83}         & 73.65            & 84.85           \\ \hline
\multirow{2}{*}{\begin{tabular}[c]{@{}c@{}}Gender \\ Recognition\end{tabular}}        & ResNet50                & \multicolumn{1}{l|}{84.06, 83.18}                                                     & 99.96 & 100   & \multicolumn{1}{c|}{87.76} & 92.62          & 59.8           & 86.51          & \multicolumn{1}{c|}{97.28}         & 66.45            & 81.75           \\
                                                                                      & ViT                     & \multicolumn{1}{l|}{85.3, 59.45}                                                      & 99.99 & 100   & \multicolumn{1}{c|}{88.57} & 93.94          & 43.79          & 87.49          & \multicolumn{1}{c|}{58.38}         & 54.81            & 53.73           \\ \hline
\multirow{2}{*}{\begin{tabular}[c]{@{}c@{}}Race\\ Recognition\end{tabular}}           & ResNet50                & 86.57, 86.44                                                                          & 99.99 & 100   & \multicolumn{1}{c|}{88.64} & 96.01          & 45.66          & 87.63          & \multicolumn{1}{c|}{97.65}         & 67.32            & 82.07           \\
                                                                                      & ViT                     & \multicolumn{1}{l|}{83.05, 86.34}                                                     & 99.87 & 100   & \multicolumn{1}{c|}{70.88} & 94.51          & 30.69          & 84.85          & \multicolumn{1}{c|}{93.39}         & 73.7             & 81.55           \\ \hline
\end{tabular}}
\label{tab:tasks}
\vspace{-0.8em}
\end{table*}

\subsection{Implementation Details}
\label{sec:Details}
For each video clip, eight face images are uniformly cropped using MTCNN~\cite{zhang2016joint} face detector. We follow~\cite{niu2020video} to extract the MSTmap with $K$=6 and $T$=300. In terms of $T$, the first 300 frames are used when the length of video $T'$ is larger than 300. If $T'$ is smaller than 300 frames, we complement the last $300-T'$ frames. $\theta$=0.8 is utilized for the weighted normalization fusion. The experiments are implemented with Pytorch on one NVIDIA V100 GPU. We use the SGD optimizer with the learning rate (lr) 0.001 and batch size 64 at the training stage. We train models with 30 epochs while lr decays with 0.1 in the 20th epoch. 
For the appearance-based unimodal experiments, we adopt ResNet50~\cite{he2016deep} and ViT-Base~\cite{dosovitskiy2020image} as our backbone models for ablation studies, and compare them with mainstream FAS (CDCN++~\cite{yu2020searching} and DeepPixel~\cite{george2019deep} with pixel-wise BCE loss) and face forgery detection methods (MesoNet~\cite{afchar2018mesonet}, Xception~\cite{rossler2019faceforensics}, and MultiAtten~\cite{zhao2021multi} with BCE loss). In terms of the rPPG-based unimodal experiments, ResNet18~\cite{niu2020video} and ViT-Base~\cite{dosovitskiy2020image} are used as our backbone models, and the classical FAS method TransRPPG~\cite{yu2021transrppg} is compared. Finally, ResNet50+ResNet18 and ViT+ViT are used as backbones for the appearance+rPPG multi-modal experiments. All ResNet and ViT initialize with ImageNet1K pretrained weights. All inputs are resized to the defaulted input sizes according to different models. For simplicity, all the joint training results are based on the fully shared backbone with 1 head 2 classes setting except the joint ViT models with name suffix `(shared 8)' only sharing the first 8 transformer blocks.

\vspace{-0.8em}
\subsection{Comparison Results on the Proposed Benchmark}
\label{sec:SOTA}


\textbf{Results of appearance-based unimodal models.} \quad  
As shown in Table~\ref{tab:appearance}, it is obvious to find that compared with separate training, most of the appearance models with joint training obtain performance gains especially for challenging cross-testing scenarios for both two tasks. Specifically, remarkable improvements can be found when joint training and cross-testing on 3DMask dataset (see the column `3DMask') on all models, i.e., 1.33\%, 11.61\%, 21.43\%, 13.88\%, 9.78\%, 20.69\%, and 27.14\% AUC improvement for MesoNet, Xception, MultiAtten, CDCN++, DeepPixel, ResNet50, and ViT, respectively.
From the perspective of testing on merged datasets (see the last bock in Table~\ref{tab:appearance}), we can find that joint training usually makes little difference (e.g., $ \textless \pm$ 4\% TPR@FPR=10\% for DeepPixel, ResNet50 and ViT) on intra-testings for both FAS and face forgery detection tasks but gives remarkable improvements on cross-testings (e.g., 9.77\%, 5.32\% and 7.49\% TPR@FPR=10\% for MultiAtten, ResNet50 and ViT on face spoofing detection while 6.78\%, 5.96\% and 6.74\% TPR@FPR=10\% for DeepPixel, ResNet50 and ViT on face forgery detection). It indicates that learning with more paired bonafide/attack data from the extra task alleviates overfitting and improves models' domain generalization capacities.

In terms of impacts of the neural architectures, we can find that generic models with stronger representation capacities (e.g., ResNet50/ViT vs. lightweight MesoNet) usually benefit more from joint training. The best cross-testing performance can be achieved by ViT(shared 8) model. We also find that the Xception model with joint training drops the cross-testing performance sharply (-7\% and -13.28\% TPR@FPR=10\% for face spoofing and forgery detection, respectively). In other words, it seems that joint training does not always enhance the generalization capacity for all model architectures. Thus, it is careful to choose or design suitable models for joint training.

\textbf{Results of rPPG-based unimodal models.} \quad  As can be seen in Table~\ref{tab:SOTArPPG}, similar to appearance-based unimodal models, joint training usually enhances rPPG models' generalization under cross-domain scenarios but makes limited improvements in intra testings. Specifically, obvious improvements can be found when joint training and cross-testing on MSU, 3DMask and DFDC datasets on all models. For instance, when separate training only on three face spoofing detection datasets recorded in lab environments, the rPPG models generalize poorly on 3DMask due to the serious unseen illumination and head movement interference. This issue is alleviated via introducing large-scale in-the-wild rPPG samples from FF++ for joint training. However, the overall performance of rPPG-based models still has a large gap with appearance-based models. The best rPPG model `ViT\_STW(shared 8)' with joint training only achieves $ \textless$40\%, $\textless$85\% and $\textless$30\% TPR@FPR=10\% for cross-domain FAS, intra-testing and cross-domain face forgery detection, respectively.

\textbf{Results of appearance+rPPG based multi-modal models.} \quad Table~\ref{tab:fusion} shows the results of appearance+rPPG models with separate training and joint training. Specifically, ResNet50/ViT-Base and ResNet18/ViT-Base are used as the backbones of appearance and rPPG modalities, respectively. On one hand, compared with separate training, joint training on all models with different fusion strategies (except ViT\_CA) improve the cross-testing performance on both tasks. Specifically, with joint training, the ResNet\_CA and ViT\_Norm\_Concat(shared 8) models generalize well (90.58\% and 88.72\% AUC) on challenging 3DMask with high-fidelity mask attacks and environmental interference, while the latter model achieves the best cross-testing FAS result, i.e., TPR=73.83\%,41.04\%@FPR=10\%,1\%. On the other hand, despite generalization capacity improvement on cross-testings with joint training, it is surprising to find from Tables~\ref{tab:appearance} and~\ref{tab:fusion} that multi-modal models with different fusion strategies achieve poorer intra- and cross-testing performance than unimodal appearance models on face forgery detection task.

\vspace{-0.6em}
\subsection{Ablation Studies}
\label{sec:Ablation}

\textbf{Efficacy of the two-branch physiological network.}\quad  
The results of rPPG-based models w/ and w/o two-branch inputs are shown in Table~\ref{tab:SOTArPPG}. Compared with models (ResNet18\_ST and ViT\_ST) with one-branch MSTmap input, models (ResNet18\_STW and ViT\_STW) with two-branch MSTmap and WaveletMap inputs achieve better performance for both intra- and cross-testings on two tasks, indicating that semantic time-frequency representation benefits the rPPG models learning more discriminative and generalized periodic features. Besides, the two-branch physiological networks are compatible with separate and joint training, consistently improving the performance compared with only MSTmap inputs.

\textbf{Efficacy of the weighted normalization fusion.}\quad  According to Equation~\ref{eq:normalization}, $\theta$ controls the contribution of the LN and BN for normalized feature fusion. As illustrated in Fig.~\ref{fig:Ablation12}(e)(f), when $\theta$=0.4 and 0.8, the multi-modal models (ResNet/ViT architectures) assembled with weighted normalization fusion generalize well for  cross-domain face spoofing and forgery detection. Based on the optimal setting with $\theta$=0.8, we can find from Table~\ref{tab:fusion} that the weighted normalization fusion consistently performs well when plugged-and-played in different architectures (e.g., ResNet and ViT). In contrast, it can be seen from the cross-testing results of `ResNet\_CA' and `ViT\_CA' that the CA fusion works well for ResNet but severely decreases the performance of ViT with joint training. We can also find from the results of `ResNet\_SE\_Concat' and `ViT\_SE\_Concat' that SE fusion is suitable for ViT but perform poorly for ResNet with joint training.

\textbf{Impacts of joint training architectures.}\quad  Table~\ref{tab:heads} shows the results of different configurations of the classification heads for joint training. We can see that when the whole backbones (ResNet and ViT) are shared, training with 1 head 2 classes (bonafide/attack) and 2 heads 2 classes settings usually perform well for the ResNet and ViT, respectively. It is interesting to find that compared with separate training, there are some negative effects with the 1 head 3 classes (bonafide, spoof, forgery) setting, indicating that the fine-grained class supervision signal might confuse the joint training. 

Besides the head settings, it is necessary to explore the impacts of shared architectures with joint training. Fig.~\ref{fig:Ablation12}(a)(b)(c)(d) illustrates the results of joint training architectures with different shared blocks using 2 heads 2 classes setting. It can be seen from Fig.~\ref{fig:Ablation12}(a)(b) that sharing all the residual blocks in ResNet has the best generalization capacity on both tasks and modalities. In terms of ViT architecture in Fig.~\ref{fig:Ablation12}(c)(d), we find that sharing the first 8 transformer blocks for joint training holds the best generalization capacity, indicating the low- and mid-level feature sharing is important for efficient joint training. In conclusion, the best joint training configuration for ResNet is to share the whole backbone with 1 head 2 classes setting while sharing the first 8 transformer blocks with 2 heads 2 classes setting for ViT.

\textbf{Impacts of sampling strategies for joint training.}\quad  The results of four sampling strategies across two tasks mentioned in Section~\ref{sec:joint} are shown in Table~\ref{tab:sampling}. It is surprising to find that utilizing the simplest sampling strategy, i.e., randomly sampling minibatch data from two tasks, performs the best in terms of modality (appearance and rPPG) and architecture (ResNet and ViT). Compared with the `Random' and `Simultaneous' sampling strategies, the `Alternating' and `Task-by-task' easily result in unstable predictions for intra- and cross-testings, which might be caused by the frequent and contradictory knowledge transfer across tasks.

\textbf{Impact of bonafide/attacks and tasks for joint training.}\quad  In previous joint training settings, bonafides and attacks from both tasks are fully exploited to learn more generalized representation. Here we explore two biased situations (i.e., only bonafides or attacks from other tasks for joint training) to see how extra bonafides or attacks contribute to the joint training. As shown in Table~\ref{tab:extraPartial}, it is interesting to find that compared with separate training, joint training with only bonafides or attacks is still able to benefit cross-testing face spoofing and forgery detection performance for both modalities except the rPPG-based `ViT\_STW' model. Note that bonafides from FAS and face forgery detection tasks are easier to collect in real-world cases, which is more practical for joint training.     

Besides the tasks related to face attack detection, it is necessary to see whether the other face analysis tasks benefit the joint training for face spoofing and forgery detection. Here we joint train models with face expression recognition (7 classes of basic emotions), gender recognition (male/female/unsure), and race recognition (Caucasian/African-American/Asian) tasks with shared 1 head multiple classes setting. The large-scale facial expression database RAF-DB~\cite{li2019reliable} is used. Table~\ref{tab:tasks} shows the results of joint training face attack detection and face expression/gender/race recognition tasks. Despite acceptable accuracy in the extra expression/gender/race recognition tasks ($\textgreater$80\%), there are limited gains for the desired face spoofing and forgery detection. Due to the low correlation among these tasks, training with more face data even drops the performance (e.g., joint training with gender recognition for ViT) compared with separate training. In other words, the benefits of joint training across face spoofing and forgery detection tasks are not only from the extra data volume but also from their high task correlations.

\begin{figure}[t]
\centering
\includegraphics[scale=0.6]{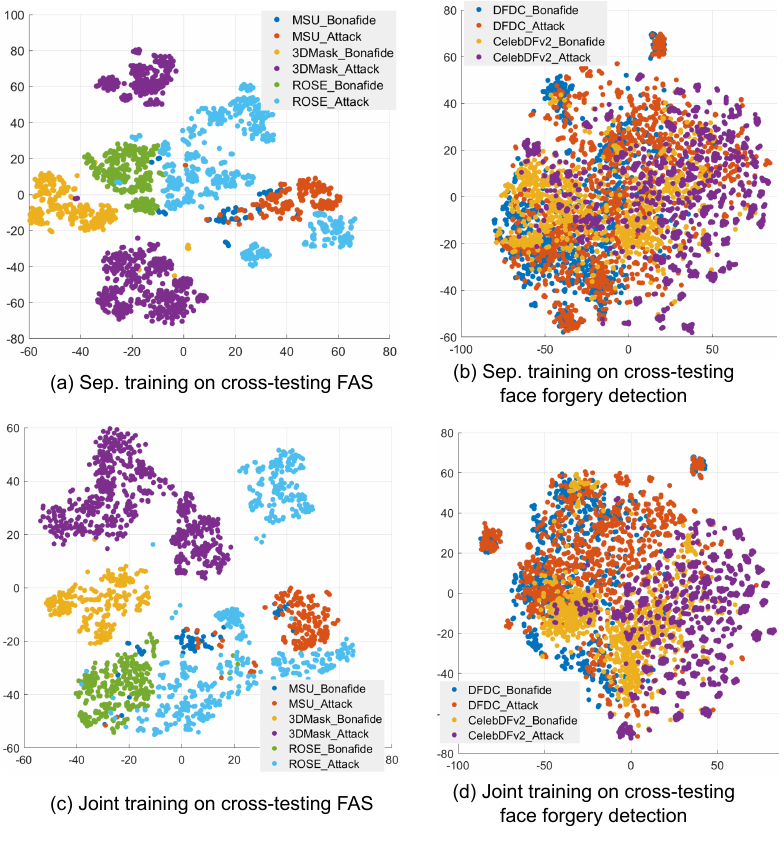}
\vspace{-0.8em}
 \caption{\small{
  Feature distribution visualization. Best view in color.}
  }
 \vspace{-0.8em}
\label{fig:visualization}
\end{figure}

\subsection{Feature Visualization}
\label{sec:Visualization2}
Distributions of the appearance feature $F_{App}$ from ViT w/ and w/o joint training are shown in Fig.~\ref{fig:visualization} via t-SNE~\cite{maaten2008visualizing}. It is clear that the features with joint training present more well-clustered
behavior than that with separate training under cross-domain scenarios. Specifically, the feature distributions of the high-fidelity face mask attacks on 3DMask (see Fig.~\ref{fig:visualization}(a) in purple) with separate training spread out on both sides while the feature distributions clusters compactly with joint training (see Fig.~\ref{fig:visualization}(c) in purple). Similarly, it is clear to see from Fig.s~\ref{fig:visualization}(b)(d) that the feature distributions of different bonafides and attacks are likely to stand alone thus easier to find clear separation hyperplanes for distinguishing bonafide/attacks (e.g., DFDC\_Bonafide in blue and CelebDFv2 in purple of Fig.~\ref{fig:visualization}(d)). 
It demonstrates that joint training across two tasks enhances the models' discrimination and generalization capacities for face attack detection.


\section{Conclusion}
\label{sec:conclusion}
In this paper, we establish the first joint face spoofing and forgery detection benchmark using both visual appearance and physiological rPPG cues. We design a two-branch physiological network for discriminative rPPG cues mining, and the weighted normalization fusion strategy for efficient appearance+rPPG multi-modal models. We also investigate prevalent deep models, fusion strategies and joint training configurations on the proposed benchmark. We note that the study of joint face spoofing and forgery detection is still at an early stage. Future directions include: 1) exploring more advanced multi-task learning and multi-modal fusion strategies; 2) besides rPPG-based color changes, exploiting semantic facial motion cues and contextual dynamic artifacts for joint face spoofing and forgery detection.


\ifCLASSOPTIONcaptionsoff
  \newpage
\fi

\bibliographystyle{IEEEtran}
\bibliography{IEEEabrv,reference}

\begin{thebibliography}{10}
\providecommand{\url}[1]{#1}
\csname url@samestyle\endcsname
\providecommand{\newblock}{\relax}
\providecommand{\bibinfo}[2]{#2}
\providecommand{\BIBentrySTDinterwordspacing}{\spaceskip=0pt\relax}
\providecommand{\BIBentryALTinterwordstretchfactor}{4}
\providecommand{\BIBentryALTinterwordspacing}{\spaceskip=\fontdimen2\font plus
\BIBentryALTinterwordstretchfactor\fontdimen3\font minus
  \fontdimen4\font\relax}
\providecommand{\BIBforeignlanguage}[2]{{%
\expandafter\ifx\csname l@#1\endcsname\relax
\typeout{** WARNING: IEEEtran.bst: No hyphenation pattern has been}%
\typeout{** loaded for the language `#1'. Using the pattern for}%
\typeout{** the default language instead.}%
\else
\language=\csname l@#1\endcsname
\fi
#2}}
\providecommand{\BIBdecl}{\relax}
\BIBdecl

\bibitem{yu2021deep}
Z.~Yu, Y.~Qin, X.~Li, C.~Zhao, Z.~Lei, and G.~Zhao, ``Deep learning for face
  anti-spoofing: A survey,'' \emph{arXiv preprint arXiv:2106.14948}, 2021.

\bibitem{galbally2014face}
J.~Galbally and S.~Marcel, ``Face anti-spoofing based on general image quality
  assessment,'' in \emph{ICPR}, 2014.

\bibitem{li2016generalized}
X.~Li, J.~Komulainen, G.~Zhao, P.-C. Yuen, and M.~Pietik{\"a}inen,
  ``Generalized face anti-spoofing by detecting pulse from face videos,'' in
  \emph{ICPR}, 2016.

\bibitem{yu2021facial}
Z.~Yu, X.~Li, and G.~Zhao, ``Facial-video-based physiological signal
  measurement: Recent advances and affective applications,'' \emph{SPM}, 2021.

\bibitem{Komulainen2014Context}
J.~Komulainen, A.~Hadid, and M.~Pietikainen, ``Context based face
  anti-spoofing,'' in \emph{BTAS}, 2013.

\bibitem{Patel2016Secure}
K.~Patel, H.~Han, and A.~K. Jain, ``Secure face unlock: Spoof detection on
  smartphones,'' \emph{TIFS}, 2016.

\bibitem{lin2019face}
B.~Lin, X.~Li, Z.~Yu, and G.~Zhao, ``Face liveness detection by rppg features
  and contextual patch-based cnn,'' in \emph{ACM ICBEA}, 2019.

\bibitem{liu20163d}
S.~Liu, P.~C. Yuen, S.~Zhang, and G.~Zhao, ``3d mask face anti-spoofing with
  remote photoplethysmography,'' in \emph{ECCV}, 2016.

\bibitem{liu2021multi}
S.-Q. Liu, X.~Lan, and P.~C. Yuen, ``Multi-channel remote photoplethysmography
  correspondence feature for 3d mask face presentation attack detection,''
  \emph{IEEE TIFS}, 2021.

\bibitem{zhang2020celeba}
Y.~Zhang, Z.~Yin, Y.~Li, G.~Yin, J.~Yan, J.~Shao, and Z.~Liu, ``Celeba-spoof:
  Large-scale face anti-spoofing dataset with rich annotations,'' in
  \emph{ECCV}, 2020.

\bibitem{liu2019deep}
Y.~Liu, J.~Stehouwer, A.~Jourabloo, and X.~Liu, ``Deep tree learning for
  zero-shot face anti-spoofing,'' in \emph{CVPR}, 2019.

\bibitem{george2019biometric}
A.~George, Z.~Mostaani, D.~Geissenbuhler, O.~Nikisins, A.~Anjos, and S.~Marcel,
  ``Biometric face presentation attack detection with multi-channel
  convolutional neural network,'' \emph{TIFS}, 2019.

\bibitem{liu2022contrastive}
A.~Liu, C.~Zhao, Z.~Yu, J.~Wan, A.~Su, X.~Liu, Z.~Tan, S.~Escalera, J.~Xing,
  Y.~Liang \emph{et~al.}, ``Contrastive context-aware learning for 3d
  high-fidelity mask face presentation attack detection,'' \emph{IEEE TIFS},
  2022.

\bibitem{yu2020searching}
Z.~Yu, C.~Zhao, Z.~Wang, Y.~Qin, Z.~Su, X.~Li, F.~Zhou, and G.~Zhao,
  ``Searching central difference convolutional networks for face
  anti-spoofing,'' in \emph{CVPR}, 2020.

\bibitem{yu2020face}
Z.~Yu, X.~Li, X.~Niu, J.~Shi, and G.~Zhao, ``Face anti-spoofing with human
  material perception,'' in \emph{ECCV}, 2020.

\bibitem{Atoum2018Face}
Y.~Atoum, Y.~Liu, A.~Jourabloo, and X.~Liu, ``Face anti-spoofing using patch
  and depth-based cnns,'' in \emph{IJCB}, 2017.

\bibitem{cai2020drl}
R.~Cai, H.~Li, S.~Wang, C.~Chen, and A.~C. Kot, ``Drl-fas: A novel framework
  based on deep reinforcement learning for face anti-spoofing,'' \emph{IEEE
  TIFS}, 2020.

\bibitem{Liu2018Learning}
Y.~Liu, A.~Jourabloo, and X.~Liu, ``Learning deep models for face
  anti-spoofing: Binary or auxiliary supervision,'' in \emph{CVPR}, 2018.

\bibitem{yu2021transrppg}
Z.~Yu, X.~Li, P.~Wang, and G.~Zhao, ``Transrppg: Remote photoplethysmography
  transformer for 3d mask face presentation attack detection,'' \emph{IEEE
  SPL}, 2021.

\bibitem{tolosana2020deepfakes}
R.~Tolosana, R.~Vera-Rodriguez, J.~Fierrez, A.~Morales, and J.~Ortega-Garcia,
  ``Deepfakes and beyond: A survey of face manipulation and fake detection,''
  \emph{Information Fusion}, 2020.

\bibitem{buchana2016simultaneous}
P.~Buchana, I.~Cazan, M.~Diaz-Granados, F.~Juefei-Xu, and M.~Savvides,
  ``Simultaneous forgery identification and localization in paintings using
  advanced correlation filters,'' in \emph{IEEE ICIP}, 2016.

\bibitem{fridrich2012rich}
J.~Fridrich and J.~Kodovsky, ``Rich models for steganalysis of digital
  images,'' \emph{IEEE TIFS}, 2012.

\bibitem{li2020face}
L.~Li, J.~Bao, T.~Zhang, H.~Yang, D.~Chen, F.~Wen, and B.~Guo, ``Face x-ray for
  more general face forgery detection,'' in \emph{CVPR}, 2020.

\bibitem{qian2020thinking}
Y.~Qian, G.~Yin, L.~Sheng, Z.~Chen, and J.~Shao, ``Thinking in frequency: Face
  forgery detection by mining frequency-aware clues,'' in \emph{ECCV}, 2020.

\bibitem{creswell2018generative}
A.~Creswell, T.~White, V.~Dumoulin, K.~Arulkumaran, B.~Sengupta, and A.~A.
  Bharath, ``Generative adversarial networks: An overview,'' \emph{IEEE SPM},
  2018.

\bibitem{ciftci2020fakecatcher}
U.~A. Ciftci, I.~Demir, and L.~Yin, ``Fakecatcher: Detection of synthetic
  portrait videos using biological signals,'' \emph{IEEE TPAMI}, 2020.

\bibitem{hernandez2022deepfakes}
J.~Hernandez-Ortega, R.~Tolosana, J.~Fierrez, and A.~Morales, ``Deepfakes
  detection based on heart rate estimation: single-and multi-frame,'' in
  \emph{Handbook of Digital Face Manipulation and Detection}, 2022.

\bibitem{qi2020deeprhythm}
H.~Qi, Q.~Guo, F.~Juefei-Xu, X.~Xie, L.~Ma, W.~Feng, Y.~Liu, and J.~Zhao,
  ``Deeprhythm: Exposing deepfakes with attentional visual heartbeat rhythms,''
  in \emph{ACM multimedia}, 2020.

\bibitem{sun2022faketransformer}
Y.~Sun, Z.~Zhang, C.~Qiu, L.~Wang, L.~Sun, and Z.~Wang, ``Faketransformer:
  Exposing face forgery from spatial-temporal representation modeled by facial
  pixel variations,'' in \emph{IEEE ICSP}, 2022.

\bibitem{ciftci2020hearts}
U.~A. Ciftci, I.~Demir, and L.~Yin, ``How do the hearts of deep fakes beat?
  deep fake source detection via interpreting residuals with biological
  signals,'' in \emph{IEEE IJCB}, 2020.

\bibitem{deb2021unified}
D.~Deb, X.~Liu, and A.~K. Jain, ``Unified detection of digital and physical
  face attacks,'' \emph{arXiv preprint arXiv:2104.02156}, 2021.

\bibitem{boulkenafet2015face}
Z.~Boulkenafet, J.~Komulainen, and A.~Hadid, ``Face anti-spoofing based on
  color texture analysis,'' in \emph{ICIP}, 2015.

\bibitem{Li2017An}
L.~Li, X.~Feng, Z.~Boulkenafet, Z.~Xia, M.~Li, and A.~Hadid, ``An original face
  anti-spoofing approach using partial convolutional neural network,'' in
  \emph{IPTA}, 2016.

\bibitem{yu2021revisiting}
Z.~Yu, X.~Li, J.~Shi, Z.~Xia, and G.~Zhao, ``Revisiting pixel-wise supervision
  for face anti-spoofing,'' \emph{IEEE TBIOM}, 2021.

\bibitem{wang2020deep}
Z.~Wang, Z.~Yu, C.~Zhao, X.~Zhu, Y.~Qin, Q.~Zhou, F.~Zhou, and Z.~Lei, ``Deep
  spatial gradient and temporal depth learning for face anti-spoofing,'' in
  \emph{CVPR}, 2020.

\bibitem{yu2021dual}
Z.~Yu, Y.~Qin, H.~Zhao, X.~Li, and G.~Zhao, ``Dual-cross central difference
  network for face anti-spoofing,'' in \emph{IJCAI}, 2021.

\bibitem{zhang2020face}
K.-Y. Zhang, T.~Yao, J.~Zhang, Y.~Tai, S.~Ding, J.~Li, F.~Huang, H.~Song, and
  L.~Ma, ``Face anti-spoofing via disentangled representation learning,'' in
  \emph{ECCV}, 2020.

\bibitem{george2019deep}
A.~George and S.~Marcel, ``Deep pixel-wise binary supervision for face
  presentation attack detection,'' in \emph{IJCB}, 2019.

\bibitem{li2018domain}
H.~Li, S.~Jialin~Pan, S.~Wang, and A.~C. Kot, ``Domain generalization with
  adversarial feature learning,'' in \emph{CVPR}, 2018.

\bibitem{jia2020single}
Y.~Jia, J.~Zhang, S.~Shan, and X.~Chen, ``Single-side domain generalization for
  face anti-spoofing,'' in \emph{CVPR}, 2020.

\bibitem{wang2020cross}
G.~Wang, H.~Han, S.~Shan, and X.~Chen, ``Cross-domain face presentation attack
  detection via multi-domain disentangled representation learning,'' in
  \emph{CVPR}, 2020.

\bibitem{liu2020disentangling}
Y.~Liu, J.~Stehouwer, and X.~Liu, ``On disentangling spoof trace for generic
  face anti-spoofing,'' in \emph{ECCV}, 2020.

\bibitem{shao2019multi}
R.~Shao, X.~Lan, J.~Li, and P.~C. Yuen, ``Multi-adversarial discriminative deep
  domain generalization for face presentation attack detection,'' in
  \emph{CVPR}, 2019.

\bibitem{shao2019regularized}
R.~Shao, X.~Lan, and P.~C. Yuen, ``Regularized fine-grained meta face
  anti-spoofing,'' in \emph{AAAI}, 2020.

\bibitem{cai2022learning}
R.~Cai, Z.~Li, R.~Wan, H.~Li, Y.~Hu, and A.~C. Kot, ``Learning meta pattern for
  face anti-spoofing,'' \emph{IEEE TIFS}, 2022.

\bibitem{li2020unseen}
Z.~Li, H.~Li, K.-Y. Lam, and A.~C. Kot, ``Unseen face presentation attack
  detection with hypersphere loss,'' in \emph{ICASSP}, 2020.

\bibitem{nikisins2018effectiveness}
O.~Nikisins, A.~Mohammadi, A.~Anjos, and S.~Marcel, ``On effectiveness of
  anomaly detection approaches against unseen presentation attacks in face
  anti-spoofing,'' in \emph{ICB}, 2018.

\bibitem{qin2021meta}
Y.~Qin, Z.~Yu, L.~Yan, Z.~Wang, C.~Zhao, and Z.~Lei, ``Meta-teacher for face
  anti-spoofing,'' \emph{IEEE TPAMI}, 2021.

\bibitem{rostami2021detection}
M.~Rostami, L.~Spinoulas, M.~Hussein, J.~Mathai, and W.~Abd-Almageed,
  ``Detection and continual learning of novel face presentation attacks,'' in
  \emph{ICCV}, 2021.

\bibitem{yu2022physformer}
Z.~Yu, Y.~Shen, J.~Shi, H.~Zhao, P.~H. Torr, and G.~Zhao, ``Physformer: facial
  video-based physiological measurement with temporal difference transformer,''
  in \emph{CVPR}, 2022.

\bibitem{yu2019remote}
Z.~Yu, W.~Peng, X.~Li, X.~Hong, and G.~Zhao, ``Remote heart rate measurement
  from highly compressed facial videos: an end-to-end deep learning solution
  with video enhancement,'' in \emph{ICCV}, 2019.

\bibitem{nowara2017ppgsecure}
E.~M. Nowara, A.~Sabharwal, and A.~Veeraraghavan, ``Ppgsecure: Biometric
  presentation attack detection using photopletysmograms,'' 2017.

\bibitem{yang2021mtd}
J.~Yang, A.~Li, S.~Xiao, W.~Lu, and X.~Gao, ``Mtd-net: learning to detect
  deepfakes images by multi-scale texture difference,'' \emph{IEEE TIFS}, 2021.

\bibitem{nguyen2019capsule}
H.~H. Nguyen, J.~Yamagishi, and I.~Echizen, ``Capsule-forensics: Using capsule
  networks to detect forged images and videos,'' in \emph{ICASSP}, 2019.

\bibitem{rossler2019faceforensics}
A.~Rossler, D.~Cozzolino, L.~Verdoliva, C.~Riess, J.~Thies, and M.~Nie{\ss}ner,
  ``Faceforensics++: Learning to detect manipulated facial images,'' in
  \emph{ICCV}, 2019.

\bibitem{chollet2017xception}
F.~Chollet, ``Xception: Deep learning with depthwise separable convolutions,''
  in \emph{CVPR}, 2017.

\bibitem{zhou2017two}
P.~Zhou, X.~Han, V.~I. Morariu, and L.~S. Davis, ``Two-stream neural networks
  for tampered face detection,'' in \emph{CVPRW}, 2017.

\bibitem{zhao2021multi}
H.~Zhao, W.~Zhou, D.~Chen, T.~Wei, W.~Zhang, and N.~Yu, ``Multi-attentional
  deepfake detection,'' in \emph{CVPR}, 2021.

\bibitem{li2021frequency}
J.~Li, H.~Xie, J.~Li, Z.~Wang, and Y.~Zhang, ``Frequency-aware discriminative
  feature learning supervised by single-center loss for face forgery
  detection,'' in \emph{CVPR}, 2021.

\bibitem{liu2021spatial}
H.~Liu, X.~Li, W.~Zhou, Y.~Chen, Y.~He, H.~Xue, W.~Zhang, and N.~Yu,
  ``Spatial-phase shallow learning: rethinking face forgery detection in
  frequency domain,'' in \emph{CVPR}, 2021.

\bibitem{hernandez2020deepfakeson}
J.~Hernandez-Ortega, R.~Tolosana, J.~Fierrez, and A.~Morales,
  ``Deepfakeson-phys: Deepfakes detection based on heart rate estimation,''
  \emph{arXiv preprint arXiv:2010.00400}, 2020.

\bibitem{caruana1997multitask}
R.~Caruana, ``Multitask learning,'' \emph{Machine learning}, 1997.

\bibitem{kokkinos2017ubernet}
I.~Kokkinos, ``Ubernet: Training a universal convolutional neural network for
  low-, mid-, and high-level vision using diverse datasets and limited
  memory,'' in \emph{CVPR}, 2017.

\bibitem{zamir2018taskonomy}
A.~R. Zamir, A.~Sax, W.~Shen, L.~J. Guibas, J.~Malik, and S.~Savarese,
  ``Taskonomy: Disentangling task transfer learning,'' in \emph{CVPR}, 2018.

\bibitem{chen2018gradnorm}
Z.~Chen, V.~Badrinarayanan, C.-Y. Lee, and A.~Rabinovich, ``Gradnorm: Gradient
  normalization for adaptive loss balancing in deep multitask networks,'' in
  \emph{ICML}, 2018.

\bibitem{kendall2018multi}
A.~Kendall, Y.~Gal, and R.~Cipolla, ``Multi-task learning using uncertainty to
  weigh losses for scene geometry and semantics,'' in \emph{CVPR}, 2018.

\bibitem{wu2020uncertainty}
J.~Wu, X.~Yu, B.~Liu, Z.~Wang, and M.~Chandraker, ``Uncertainty-aware
  physically-guided proxy tasks for unseen domain face anti-spoofing,''
  \emph{arXiv preprint arXiv:2011.14054}, 2020.

\bibitem{ming2022vitranspad}
Z.~Ming, Z.~Yu, M.~Al-Ghadi, M.~Visani, M.~MuzzamilLuqman, and J.-C. Burie,
  ``Vitranspad: Video transformer using convolution and self-attention for face
  presentation attack detection,'' in \emph{ICIP}, 2022.

\bibitem{niu2020video}
X.~Niu, Z.~Yu, H.~Han, X.~Li, S.~Shan, and G.~Zhao, ``Video-based remote
  physiological measurement via cross-verified feature disentangling,'' in
  \emph{ECCV}, 2020.

\bibitem{yu2020multi}
Z.~Yu, Y.~Qin, X.~Li, Z.~Wang, C.~Zhao, Z.~Lei, and G.~Zhao, ``Multi-modal face
  anti-spoofing based on central difference networks,'' in \emph{CVPRW}, 2020.

\bibitem{ioffe2015batch}
S.~Ioffe and C.~Szegedy, ``Batch normalization: Accelerating deep network
  training by reducing internal covariate shift,'' in \emph{ICML}, 2015.

\bibitem{ba2016layer}
J.~L. Ba, J.~R. Kiros, and G.~E. Hinton, ``Layer normalization,'' \emph{arXiv
  preprint arXiv:1607.06450}, 2016.

\bibitem{hu2018squeeze}
J.~Hu, L.~Shen, and G.~Sun, ``Squeeze-and-excitation networks,'' in
  \emph{CVPR}, 2018.

\bibitem{casiasurf}
S.~Zhang, X.~Wang, A.~Liu, C.~Zhao, J.~Wan, S.~Escalera, H.~Shi, Z.~Wang, and
  S.~Z. Li, ``A dataset and benchmark for large-scale multi-modal face
  anti-spoofing,'' in \emph{CVPR}, 2019.

\bibitem{yu2022flexible}
Z.~Yu, C.~Zhao, K.~H. Cheng, X.~Cheng, and G.~Zhao, ``Flexible-modal face
  anti-spoofing: A benchmark,'' \emph{arXiv preprint arXiv:2202.08192}, 2022.

\bibitem{erdogmus2014spoofing}
N.~Erdogmus and S.~Marcel, ``Spoofing face recognition with 3d masks,''
  \emph{IEEE TIFS}, 2014.

\bibitem{wen2015face}
D.~Wen, H.~Han, and A.~K. Jain, ``Face spoof detection with image distortion
  analysis,'' \emph{IEEE TIFS}, 2015.

\bibitem{yu2020fas2}
Z.~{Yu}, J.~{Wan}, Y.~{Qin}, X.~{Li}, S.~Z. {Li}, and G.~{Zhao}, ``Nas-fas:
  Static-dynamic central difference network search for face anti-spoofing,''
  \emph{IEEE TPAMI}, 2020.

\bibitem{li2018unsupervised}
H.~Li, W.~Li, H.~Cao, S.~Wang, F.~Huang, and A.~C. Kot, ``Unsupervised domain
  adaptation for face anti-spoofing,'' \emph{IEEE TIFS}, 2018.

\bibitem{dolhansky2019deepfake}
B.~Dolhansky, R.~Howes, B.~Pflaum, N.~Baram, and C.~C. Ferrer, ``The deepfake
  detection challenge (dfdc) preview dataset,'' \emph{arXiv preprint
  arXiv:1910.08854}, 2019.

\bibitem{li2020celeb}
Y.~Li, X.~Yang, P.~Sun, H.~Qi, and S.~Lyu, ``Celeb-df: A large-scale
  challenging dataset for deepfake forensics,'' in \emph{CVPR}, 2020.

\bibitem{ACER}
international organization~for standardization, ``Iso/iec jtc 1/sc 37
  biometrics: Information technology biometric presentation attack detection
  part 1: Framework.'' in \emph{https://www.iso.org/obp/ui/iso}, 2016.

\bibitem{wang2022domain}
Z.~Wang, Z.~Wang, Z.~Yu, W.~Deng, J.~Li, T.~Gao, and Z.~Wang, ``Domain
  generalization via shuffled style assembly for face anti-spoofing,'' in
  \emph{CVPR}, 2022.

\bibitem{li2019reliable}
S.~Li and W.~Deng, ``Reliable crowdsourcing and deep locality-preserving
  learning for unconstrained facial expression recognition,'' \emph{IEEE TIP},
  2019.

\bibitem{zhang2016joint}
K.~Zhang, Z.~Zhang, Z.~Li, and Y.~Qiao, ``Joint face detection and alignment
  using multitask cascaded convolutional networks,'' \emph{IEEE SPL}, 2016.

\bibitem{he2016deep}
K.~He, X.~Zhang, S.~Ren, and J.~Sun, ``Deep residual learning for image
  recognition,'' in \emph{CVPR}, 2016.

\bibitem{dosovitskiy2020image}
A.~Dosovitskiy, L.~Beyer, A.~Kolesnikov, D.~Weissenborn, X.~Zhai,
  T.~Unterthiner, M.~Dehghani, M.~Minderer, G.~Heigold, S.~Gelly \emph{et~al.},
  ``An image is worth 16x16 words: Transformers for image recognition at
  scale,'' in \emph{ICLR}, 2021.

\bibitem{afchar2018mesonet}
D.~Afchar, V.~Nozick, J.~Yamagishi, and I.~Echizen, ``Mesonet: a compact facial
  video forgery detection network,'' in \emph{WIFS}, 2018.

\bibitem{maaten2008visualizing}
L.~v.~d. Maaten and G.~Hinton, ``Visualizing data using t-sne,'' \emph{JMLR},
  2008.

\end{thebibliography}

\end{document}